\pdfoutput=1
\pdfoutput=1
\pdfoutput=1
\pdfoutput=1
\pdfoutput=1
\documentclass[journal,draftcls,onecolumn,12pt,twoside]{IEEEtranTCOM}
\normalsize

\renewcommand{\baselinestretch}{1.4}
\usepackage{enumerate}
\usepackage{color}

\usepackage{cuted}
\usepackage{cite}

\usepackage{float}
\newfloat{figtab}{htb}{fgtb}
\makeatletter
\newcommand\figcaption{\def\@captype{figure}\caption}
\newcommand\tabcaption{\def\@captype{table}\caption}
\makeatother

\usepackage{graphicx}
\usepackage{booktabs}
\usepackage{algorithm}
\usepackage{bm}
\ifCLASSINFOpdf
\else
\fi

\usepackage{amsthm}
\usepackage{amssymb,amsfonts}
\usepackage{amsfonts,balance}

\usepackage[cmex10]{amsmath}
\usepackage{lipsum}

\newtheorem{theorem}{Theorem}

\usepackage{algorithmic}

\usepackage{array}
\usepackage{multirow,makecell}
\usepackage[caption=false,font=footnotesize]{subfig}

\usepackage{verbatim}

\begin{document}
%
\title{Adaptive Federated Pruning in Hierarchical Wireless Networks}
%
%
\author{Xiaonan~Liu,  Shiqiang~Wang,~\IEEEmembership{Member,~IEEE,} Yansha Deng,~\IEEEmembership{Senior~Member,~IEEE,} Arumugam~Nallanathan,~\IEEEmembership{Fellow,~IEEE}
\thanks{X. Liu and A. Nallanathan are with the School of Electronic Engineering and Computer Science, Queen Mary University of London (QMUL), U.K. (e-mail: \{x.l.liu, a.nallanathan\}@qmul.ac.uk).}
\thanks{Y. Deng is with the Department of Engineering, King’s College London, London, WC2R 2LS, U.K. (e-mail: yansha.deng@kcl.ac.uk). (Corresponding author: Yansha Deng).}
\thanks{S. Wang is with IBM T. J. Watson Research Center, NY, USA. (e-mail: wangshiq@us.ibm.com).}
}

\maketitle

\begin{abstract}
Federated Learning (FL) is a promising privacy-preserving distributed learning framework where a server aggregates models updated by multiple devices without accessing their private datasets. Hierarchical FL (HFL), as a device-edge-cloud aggregation hierarchy, can enjoy both the cloud server's access to more datasets and the edge servers' efficient communications with devices. However, the learning latency increases with the HFL network scale due to the increasing number of edge servers and devices with limited local computation capability and communication bandwidth. To address this issue, in this paper, we introduce model pruning for HFL in wireless networks to reduce the neural network scale. We present the convergence analysis of an upper on the $l_2$-norm of gradients for HFL with model pruning, analyze the computation and communication latency of the proposed model pruning scheme, and formulate an optimization problem to maximize the convergence rate under a given latency threshold by jointly optimizing the pruning ratio and wireless resource allocation. By decoupling the optimization problem and using Karush–Kuhn–Tucker (KKT) conditions, closed-form solutions of pruning ratio and wireless resource allocation are derived. Simulation results show that our proposed HFL with model pruning achieves similar learning accuracy compared with the HFL without model pruning and reduces about $50\%$ communication cost.
\end{abstract}




%
\IEEEpeerreviewmaketitle


\section{Introduction}
In recent years, with the availability of enormous mobile data and growing privacy concerns, stringent privacy protection laws, such as the European Commission’s General Data Protection Regulation (GDPR) \cite{custers} and the Consumer Privacy Bill of Rights in the U.S. \cite{Gaff}, have been proposed. These potentially impede the development of Artificial Intelligence (AI)-based frameworks which are mainly cloud/edge-centric, where data is delivered to a cloud/edge server for data analysis by machine learning (ML) algorithms \cite{Rimal,pingli}.

In response, Federated Learning (FL) has emerged as a powerful privacy-preserving distributed ML architecture  \cite{B_McMahan}. The standard steps of FL are: 1) each device uses its dataset to train a local model; 2) devices send their local models to the server for model aggregation; 3) the server transmits the updated global model back to devices. These steps are repeated across multiple iterations until convergence. Therefore, in FL, only the updated local models, rather than the raw data, are transmitted to the server for model aggregation, which enables privacy preservation towards the development of AI-empowered applications.

However, FL usually suffers from a bottleneck of communication overhead before achieving convergence because of long transmission latency between the cloud server and devices \cite{Saxena}. Meanwhile, the wireless channel between the cloud server and devices can be unreliable due to wireless fading, which further affects model sharing and degrades learning performance under latency constraints. In addition, since some ML models have large size, directly communicating with the cloud server over the wireless channel by a massive number of devices could lead to congestion in the backbone network.

To mitigate this issue, hierarchical federated learning (HFL) framework is proposed, where small-cell base stations are equipped with edge servers to perform edge aggregation of local models from local devices \cite{HFEL1}. When edge servers achieve a certain learning accuracy, updated edge models are transmitted to the cloud server for global aggregation. Therefore, by leveraging edge servers as intermediaries to perform partial model aggregation in proximity, the communication overhead reduces significantly. Also, more efficient communication and computation resource allocation, such as energy and bandwidth allocation, can be achieved by the coordination of edge servers \cite{WYBLim,Kar}.

Unfortunately, there are still challenges for HFL in wireless networks. First, with the increasing number of edge servers and devices, the limited wireless resources cannot provide efficient services, which leads to high communication latency for model uploading. Second, the computation capabilities of devices are limited, which results in high computation latency, especially for large-scale learning models. To address these issues, federated pruning is introduced in \cite{9598845,9762360}, where the model size is adapted during FL to reduce both communication and computation overhead and minimize the overall training time, while maintaining a similar learning accuracy as the original model. However, authors in \cite{9598845,9762360} only considered one edge server and multiple devices so that the number of devices access to the edge server is limited, leading to inevitable training performance loss \cite{9148862}. Therefore, we see a necessity for leveraging a cloud server to access the massive training samples, while edge servers enjoy quick model updates from their local devices with model pruning. 

Motivated by the above, in this work, we propose a joint model pruning and wireless resource allocation for HFL in wireless networks. First, the variation of computation and communication latency caused by the model pruning are mathematically analyzed. Second, the convergence analysis of HFL with model pruning is proposed. Then, the pruning ratio and wireless resource allocation under latency and bandwidth constraints are jointly optimized to improve learning performance. The main contributions are summarized as follows.
\begin{itemize}
    \item To adapt to dynamical wireless environments, we propose HFL with adaptive model pruning for device-edge-cloud wireless networks. To the best of our knowledge, this is the first paper considering adaptive model pruning in HFL.
    \item We model the number of model weights based on the pruning ratio. Also, we model the computation and communication latency of the HFL framework under a given pruning ratio. Furthermore, we analyze the convergence of an upper bound on the $l_2$-norm of gradients for HFL with adaptive model pruning. Then, the pruning ratio and wireless resource allocation are jointly optimized to minimize the upper bound under latency and bandwidth constraints.
    \item To obtain the optimal closed-form solutions of pruning ratio and wireless resource allocation in each communcation round, we decouple the optimization problem into two sub-problems and deploy Karush–Kuhn–Tucker (KKT) conditions.
    \item Simulation results show that our proposed HFL with adaptive model pruning achieves similar learning accuracy compared to the HFL without model pruning and decreases about $50 \%$ communication cost. Also, the learning accuracy of our proposed HFL with adaptive model pruning is larger than that of non-hierarchy model pruning.
\end{itemize}

The rest of this paper is organized as follows. Section II presents the related works. The system model is detailed in Section III. The convergence analysis and problem formulation are presented in Section IV. The optimal pruning ratio and wireless resource allocation are described in Section V. The simulation results and conclusions are detailed in Section VI and Section VII, respectively.

\section{Related Works}
In this section, related works on neural network pruning, efficient FL, and resource allocation and device selection in HFL are briefly introduced in the following three subsections.

\subsection{Neural Network Pruning}
To reduce the complexity of neural networks, importance-based pruning has become popular in recent years \cite{P_Molchanov}, where weights with smaller importance are removed from the network. It is observed that directly training the pruned network can reach a similar accuracy as pruning a pre-trained original network. In addition to the importance-based pruning that trains the learning model until convergence before the next pruning step, there are iterative pruning methods where the model is pruned after every few steps of training \cite{9878734}. Furthermore, a dynamic pruning approach that allows the neural network to grow and shrink during training was proposed in \cite{Lin2020Dynamic}. However, the pruning techniques in \cite{P_Molchanov,9878734,Lin2020Dynamic} are mainly considered in centralized learning with full access to training data, which is fundamentally different from our adaptive HFL pruning that works with distributed datasets at local devices and preserves device privacy.  

\subsection{Efficient Federated Learning}
To improve the computation and communication efficiency of FL, federated dropout (FedDrop) was studied \cite{9707474,9484526,Cheng_2022_CVPR,NEURIPS2021_6aed000a}. A FedDrop scheme in \cite{9707474} was proposed building on the classic dropout scheme for random model pruning. Specifically, in each iteration of the FL algorithm, several subnets were independently generated from the global model at the server using heterogeneous dropout rates, each of which was adapted to the state of an assigned channel. An adaptive FedDrop technique was proposed in \cite{9484526} to optimize both server-device communication and computation costs by allowing devices to train locally on a selected subset of the global model. In \cite{Cheng_2022_CVPR}, the authors argued that the metrics used to measure the performance of FedDrop and its variants were misleading, and they proposed and performed new experiments which suggested that FedDrop was actually detrimental to scaling efforts. In \cite{NEURIPS2021_6aed000a}, ordered FedDrop was introduced, where a mechanism that achieved an ordered, nested representation of knowledge in neural networks and enabled the extraction of lower footprint submodels without the need for retraining. FedDrop is a simple way to prevent the learning model from overfitting through randomly dropping neurons and is only used during the training phase, which decreases communication and computation latencies and slightly improves learning accuracy. However, during the testing phase, the whole learning model is transmitted between the server and devices, which cannot guarantee efficient FL.

To address the issue in FedDrop, federated pruning was proposed \cite{9598845,9762360}. In \cite{9598845}, model pruning for wireless FL was introduced to reduce the neural network scale, and device selection was also considered to further improve the learning performance.  By removing the stragglers with low computing power or poor channel condition, the model aggregation loss caused by model pruning could be alleviated and the communication overhead could be effectively reduced. In \cite{9762360}, a novel FL approach with adaptive and distributed parameter pruning, called PruneFL, was proposed, which adapted the model size during FL to reduce both communication and computation overhead
and minimized the overall training time, while maintaining a similar accuracy as the original model. PruneFL included initial pruning at a selected device and further pruning as part of the
FL process. The model size was adapted during this process, which included maximizing the approximate empirical risk reduction divided by the time of one FL round. However, authors in \cite{9598845,9762360} only considered one edge server and multiple devices, the performance of federated model pruning in hierarchical wireless networks was still unclear. 

Except from FedDrop and federated pruning for efficient FL in training or testing phases, some other methods such as device-to-device (D2D)-assisted efficient FL protocol, were proposed to guarantee communication-efficient FL in wireless networks \cite{RA14,RA15,RA16}. In \cite{RA14}, a D2D-assisted FL scheme, called (D2D-FedAvg), over mobile edge computing (MEC) networks was proposed to minimize the communication cost. D2D-FedAvg created a two-tier learning model where D2D learning groups communicated their results as a single entity to the MEC for traffic reduction. Also, D2D grouping, master device selection, and D2D exit were proposed to form a complete D2D-assisted FedAvg. In \cite{RA15}, sign-SGD was considered in a D2D-assisted FL scheme to minimize communication costs. In \cite{RA16}, an efficient FL protocol called local-area network (LAN) FL was proposed, which involved a hierarchical aggregation mechanism in LAN due to its abundant
bandwidth and almost negligible monetary cost than wide-area network (WAN). LAN FL could accelerate the learning process and reduce the monetary cost with frequent local aggregation in the same LAN and infrequent global aggregation on a cloud across WAN. However, the computation efficiency of FL in \cite{RA14,RA15,RA16} was not considered, especially for devices with limited computation capability.

\subsection{Resource Allocation and Device Selection}
In \cite{RA1,RA2,RA3,RA4,RA5,RA8}, computation and communication resource allocation and edge association of HFL were investigated. Specifically, in \cite{RA1}, the sum of system and learning costs was minimized by optimizing bandwidth, computing frequency, power allocation, and sub-carrier assignment by  successive convex approximation and Hungarian algorithms. In \cite{RA2}, a conflict graph-based solution was proposed to minimize the overall energy consumption of training local models subject to HFL latency constraints. In \cite{RA3}, a hierarchical game framework was proposed to study the dynamics of edge association and resource allocation in self-organizing HFL networks, and a Stackelberg differential game was used to model the optimal bandwidth and reward allocation strategies of the edge servers and devices. In \cite{RA4}, a fog-enabled FL framework was proposed to facilitate distributed learning for delay-sensitive applications in resource-constrained internet of things environments, where a greedy heuristic approach was formulated to select an optimal fog node for model aggregation. In \cite{RA5}, a multi-layer FL protocol, called HybridFL, was designed for a MEC architecture, which could mitigate stragglers and end device drop-out. In \cite{RA8}, an in-network aggregation process (INA) was designed to enable decentralizing the model aggregation process at the server, thereby minimizing the training latency for the whole FL network. However, the whole FL learning model still needs to be transmitted between servers and devices in  \cite{RA1,RA2,RA3,RA4,RA5,RA8}, which cannot guarantee computation and communication efficient HFL.

\begin{figure*}[!h]
    \centering
    \includegraphics[width=5.0 in]{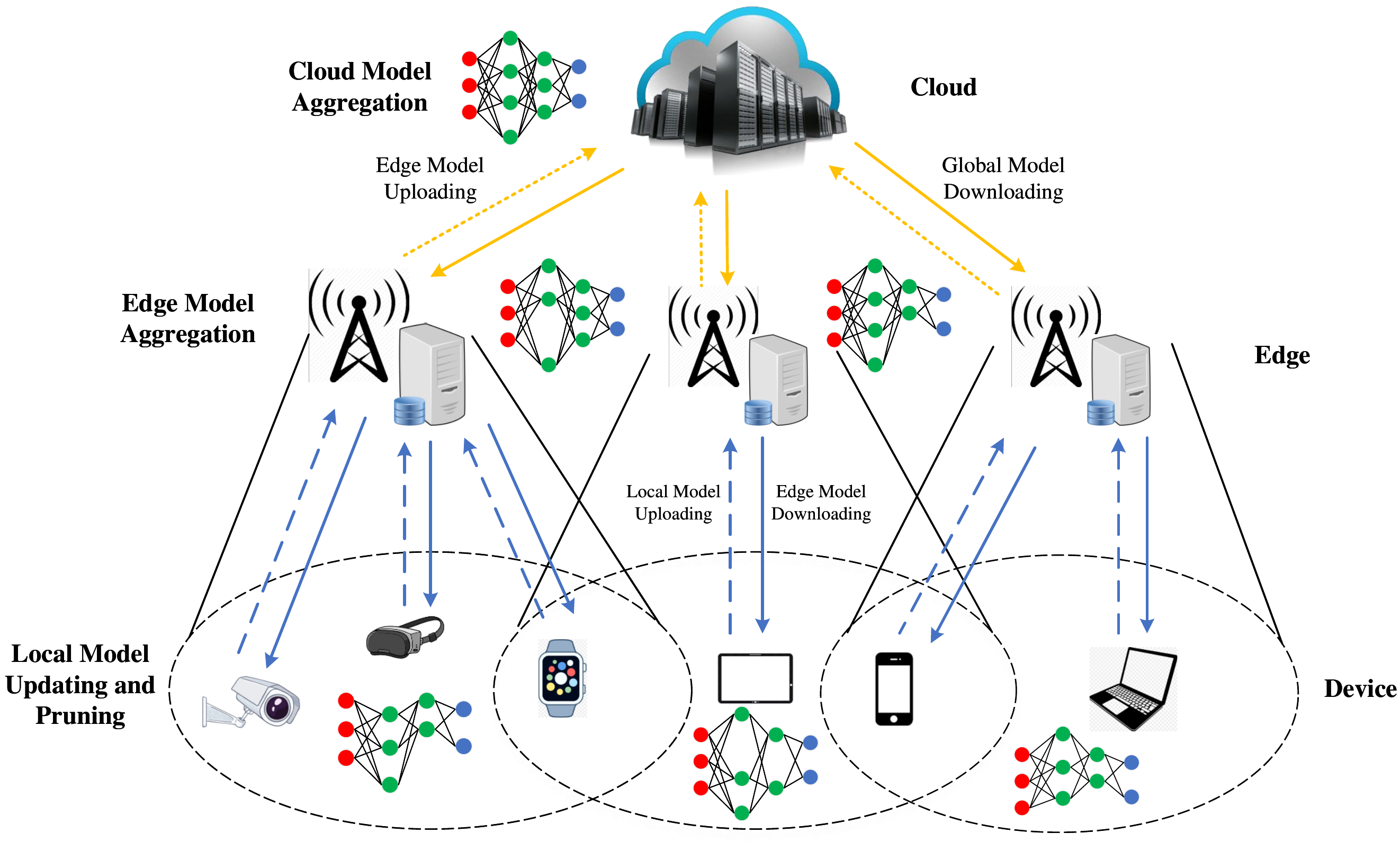}
    \caption{Hierarchical Federated Learning (HFL) framework.}\vspace{-10mm}
    \label{basic_modules}
\end{figure*}

\section{System Model}
In a HFL network, we assume a set of edge servers $\mathcal{K} = \{k = 1, 2,..., K\}$, a set of mobile devices $\mathcal{N} = \{n = 1, 2,..., N\}$, and a cloud server $S$. The $k$th edge server provides wireless connections for $\mathcal{N}_k\in\mathcal{N}$ mobile devices and is connected to the cloud server $S$ through a fiber link. Each edge server is equipped with $M$ antennas and each mobile device is equipped with a single antenna. In addition, the $n$th device has a local dataset $\mathcal{D}_n = \{(\bm{x}_i, y_i)\}_{i = 1}^{D_n}$, where $\bm{x}_i$ is the $i$th input data sample, $y_i$ is the corresponding labeled output of $\bm{x}_i$, and $D_n$ is the number of data samples.

\subsection{Model Pruning}
In the HFL framework, the scale of neural networks can be very large with increasing requirements of learning performance, such as high learning accuracy. Consequently, model updating in local devices and transmission between edge servers and devices could cause high computation and communicatioin latency. To solve these problems, model pruning is deployed to decrease the model size.

Pruning unimportant neurons or weights effectively decreases the model size and only causes a small performance loss. The learning accuracy only decreases dramatically with a high pruning ratio. According to \cite{P_Molchanov}, the importance of weight is quantified by the error induced by removing it, and the induced error is measured as a squared difference of prediction errors with and without the $j$th weight $w_{n,j}$ of the $n$th device, which is denoted as
\begin{equation}
    {\mathcal{I}}_{n,j} = \left(F_{n}(\bm{w}_n) - F_{n}(\bm{w}_n | w_{n,j} = 0)\right)^2,
\end{equation}
where $F_{n}(\bm{w}_n)$ and $\bm{w}_n$ are the local loss function and local model of the $n$th device, respectively.
The larger the error is, the more important the weight will be. However, calculating ${\mathcal{I}}_{n,j}$ for each weight of the $n$th device in (1) is computationally expensive, especially when the $n$th device has a large number of model weights. To decrease the computational complexity of the importance calculation, we calculate the difference between the $j$th local model weight and the updated $j$th local model weight as
\begin{equation}
    \hat{{\mathcal{I}}}_{n,j} = |{w}_{n,j} - \hat{w}_{n,j}|.
\end{equation}
The importance calculation in (2) is easily computed since the updated local model weight $\hat{w}_{n,j}$ is already available from backpropagation.

The primary objective of model pruning is to alleviate the high computational demands of the training and inference phases. When the $l$th layer of the learning model is pruned through importance-based model pruning given pruning ratio $\rho_{n,l}$, there is no need to perform forward and backward passes or gradient updates on the pruned units. As a result, model pruning offers gains both in terms of floating point operation (FLOP) count and model size \cite{Horvath}. Specifically, for the $l$th fully-connected layer, the number of weights is calculated as
\begin{equation}
    W_{n,l} = \lceil\rho_{n,l}W_{n,l,\text{in}}\rceil\lceil\rho_nW_{n,l,\text{out}}\rceil,
\end{equation}
where $W_{n,l,\text{in}}$ and $W_{n,l,\text{out}}$ correspond to the number of input and output weights, respectively, and the number of weights is decreased by $\frac{W_{n,l,\text{in}}W_{n,l,\text{out}}}{\lceil\rho_{n,l}W_{n,l,\text{in}}\rceil\lceil\rho_{n,l}W_{n,l,\text{out}}\rceil}\sim\frac{1}{\rho_{n,l}^\text{out}}$. Furthermore, the bias terms are reduced by a factor of $\frac{W_{n,l,\text{out}}}{\lceil\rho_{n,l}W_{n,l,\text{out}}\rceil}\sim\frac{1}{\rho_{n,l}}$. For simplicity, we directly deploy $\rho_{n}$ to denote the pruning ratio of the $n$th device in the following sections.

\subsection{Learning Process in HFL}
The proposed HFL architecture is shown in Fig. 1, where learning models are aggregated in the edge and cloud servers. As a result, learning models updated by mobile devices in a global iteration include edge aggregation and cloud aggregation. To quantify training overhead in the HFL framework, we formulate latency overhead in edge and cloud aggregation within one global iteration. The learning process is introduced as follows:

\subsubsection{Edge Aggregation} This stage has five steps, including cloud model broadcasting, edge model broadcasting, local model updating, transmission, and aggregation. That is, the cloud server broadcasts the cloud model to edge servers, each edge server broadcasts the edge model to its associated mobile devices, and mobile devices update local models with their datasets and transmit them to their associated edge servers for model aggregation, which are introduced as following steps.

\paragraph{Step 1. Cloud Model Broadcast} In the $q$th global communication round, the cloud model $\bm{w}_{G}^{q}$ is broadcast to all edge servers. Considering a convolutional neural network (CNN) and the number of weights $W$ in CNN is calculated as
\begin{align}
    W = W_{\text{conv}} + W_{\text{fully}}= W_{\text{conv}} + \sum_{l=1}^{L-1}N_{l}N_{l+1},
\end{align}
where $W_{\text{conv}}$ is the number of weights of convolutional layers, $L$ is the number of fully-connected layers, and $\sum_{l=1}^{L-1}N_{l}N_{l+1}$ is the total number of weights of fully-connected layers. Since the fiber links between the cloud and edge servers have a high transmission rate, the transmission latency between them is ignored.

\paragraph{Step 2. Edge Model Broadcasting} The $k$th edge server transmits the received cloud model $\bm{w}_{G}^{q}$ to its associated mobile devices through downlink transmission. In actual scenarios, the transmission latency of downlink weights broadcast is very small due to sufficient downlink broadcast channel bandwidth. Therefore, downlink transmission latency is ignored in the study of this paper.

\paragraph{Step 3. Local Model Updating} When the $n$th device receives the model from the $k$th edge server in the beginning of the $e$th edge communication round, it deploys a pruning mask $\bm{m}_{k,n}^{q,e}$ to prune the received edge model $\bm{w}_{k,n}^{q,e}$, which is calculated as
\begin{equation}
    \bm{w}_{k,n}^{q,e,0} = \bm{w}_{k,n}^{q,e}\odot\bm{m}_{k,n}^{q,e}.
\end{equation}
In the pruning mask $\bm{m}_{k,n}^{q,e}$, if ${m}_{k,n}^{q,e,j} = 1$, $\bm{w}_{k,n}^{q,e,0}$ contains the $j$th model weight, otherwise, ${m}_{k,n}^{q,e,j} = 0$, and ${m}_{k,n}^{q,e,j}$ is determined by (2). In addition, $t$ means the $t$th iteration in local model updating. Then, by gradient descent, the $n$th device updates the pruned local model $\bm{w}_{k,n}^{q,e,0}$. Given a pruning ratio $\rho_{n, e}$ of the $n$th mobile device, the number of weights after pruning is calculated as
\begin{equation}
    W_{\rho_{n, e}} = W_{n,\text{conv}} + (1 - \rho_{n, e})W_{n,\text{fully}}.
\end{equation}
In (6), we mainly consider weight pruning in the fully-connected layer rather than the convolutional layer. It is because pruning in the convolutional layer decreases the robust capability of CNN.

The local loss $F_{n}(\bm{w}_{k,n}^{q,e,t})$ of the $n$th device in the $t$th iteration is defined on its local dataset $\mathcal{D}_n$ and is denoted as
\begin{equation}
    F_n(\bm{w}_{k,n}^{q,e,t}) = \frac{1}{D_n}\sum_{i=1}^{D_n} f_n(\bm{x}_i, y_i, \bm{w}_{k,n}^{q,e,t}),
\end{equation}
where $f_n(\bm{x}_i, y_i, \bm{w}_{k,n}^{q,e,t})$ is the loss function (e.g., cross-entropy and mean square error (MSE)) that denotes the difference between the model output and the desired output based on the local model $\bm{w}_{k,n}^{q,e,t}$. The fact that calculating the
loss over the whole dataset is time-consuming, and in some cases it is not feasible because of the limited memory capacity of the mobile device, we employ minibatch stochastic gradient descent (SGD) in which the $n$th mobile device deploys a sub-dataset  of its dataset to calculate the loss. The local model updating in the $t$th iteration is calculated as
\begin{align}
    \bm{w}_{k,n}^{q,e,t+1} &= \bm{w}_{k,n}^{q,e,t} - \eta\nabla F_{n}(\bm{w}_{k,n}^{q,e,t}, \xi_{k,n}^{q,e,t})\odot\bm{m}_{k,n}^{q,e},
\end{align}
where $\nabla F_{n}(\bm{w}_{k,n}^{q,e,t}, \xi_{k,n}^{q,e,t})$ is the gradient in the $t$th iteration, $\eta$ is the learning rate, $\xi_{k,n}^{q,e,t}\subseteq\mathcal{D}_{n}$ is the mini-batch randomly selected from the data samples $\mathcal{D}_{n}$ of the $n$th mobile device.

Then, we can calculate the computation latency incurred by the $n$th device. We assume that the number of CPU cycles for the $n$th device to update one model weight is $C_n$, thus, the total number of CPU cycles to run one local iteration is $C_nW_{\rho_{n, e}}$. We denote that the allocated CPU frequency of the $n$th device for computation is $f_{n}$ with $f_{n}\in[f_n^{\min}, f_n^{\max}]$. Therefore, the total latency of local iterations is calculated as
\begin{equation}
    T_{n, e}^{\text{cmp}} = \frac{TC_nW_{\rho_{n, e}}}{f_{n}}.
\end{equation}
where $T$ is the number of iterations.

\paragraph{Step 4. Local Model Uplink Transmission} After finishing local model updating, the $n$th device transmits its updated local model ${\bm{w}}_{k,n}^{q,e,T}$ to the $k$th edge server, which results in wireless transmission latency. We assume that the set of mobile devices associated with the $k$th server is $\mathcal{S}_k$ with $\mathcal{S}_k\subseteq\mathcal{N}$.

The achievable transmission rate between the $n$th mobile device and the $k$th edge server in the $e$th edge communication round is denoted as
\begin{equation}
    R_{n,k,e}^{\text{up}} = {b}_{n,e}B\log_{2}\left(1 + \frac{g_{n,k}^{e}p_n}{\sigma^2}\right),
\end{equation}
where ${b}_{n,e}$ is the bandwidth fraction allocated to the $n$th mobile device in the $e$th edge communication round, $B$ is the total bandwidth allocated to each edge server, $g_{n,k}^{e}$ is the channel gain between the $n$th mobile device and the $k$th edge server, $p_n$ is the transmission power of the $n$th mobile device, and $\sigma^2$ is the noise power. Then, the uplink transmission latency from the $n$th mobile device to the $k$th edge server is calculated as
\begin{equation}
    T_{n,k,e}^{\text{up}} = \frac{\hat{q}W_{\rho_{n,e}}}{R_{n,k,e}^{\text{up}}},
\end{equation}
where $\hat{q}$ is the quantization bit.

\paragraph{Step 5. Edge Model Aggregation} Because of model pruning, some model weights are not contained in the received local models. Let $\mathcal{N}_{k,q,e}^{j}$ be the set of mobile devices associated with the $k$th edge server and containing the $j$th model weight in the $e$th edge communication round. Then, edge model update of the $j$th model weight is performed by aggregating local models with the $j$th model weight available, which is calculated as
\begin{equation}
    \hat{w}_{k}^{q,e,T,j} = \frac{1}{|\mathcal{N}_{k,q,e}^{j}|}\sum_{n\in\mathcal{N}_{k,q,e}^{j}}{w}_{k,n}^{q,e,T,j},
\end{equation}
where $|\mathcal{N}_{k,q,e}^{j}|$ is the number of local models containing the $j$th model weight.

Then, the $k$th edge server delivers ${\bm{w}}_{k}^{q,e+1}$ to its associated mobile devices in $\mathcal{S}_k$ for the next round of local model updating in step 1. Actually, steps 1 to 5 of edge aggregation continue iterating until the $k$th edge server reaches a certain level of accuracy. Each edge server does not access the local dataset of each mobile device, which preserves personal data privacy. Since each edge server typically has high computation capability, the computation latency of edge model aggregation is neglected.

\subsubsection{Cloud Aggregation} This stage has two steps, including edge model uploading and cloud model aggregation. First, the $k$th edge server transmits updated ${\bm{w}}_{k}^{q,E}$ to the cloud for global aggregation after $E$ edge communication rounds. Then, the cloud server aggregates edge models from all edge servers as
\begin{equation}
    \bm{w}_{G}^{q+1} = \frac{1}{|\mathcal{K}|}\sum_{k\in\mathcal{K}}{\bm{w}}_{k}^{q,E},
\end{equation}
where $|\mathcal{K}|$ is the number of edge servers. The detailed HFL with model pruning is presented in $\textbf{Algorithm 1}$.

\begin{algorithm}[t]
\begin{algorithmic}[1]
\caption{HFL with model pruning}
\STATE Local dataset $\mathcal{D}_n$ on $N_k$ local devices associated with the $k$th edge server, learning rate $\eta$, pruning policy $\mathbb{P}$, number of local epochs $T$, number of edge communication rounds $E$, edge model parameterized by $\bm{w}_{k}^{q,e}$, number of global communication rounds $Q$, global model parameterized by $\bm{w}_{G}^{q}$.
\FOR{global communication round $q$ = 1,...,$Q$}
\STATE Generate $\bm{w}_{k}^{q,0} = \bm{w}_{G}^{q}$.
\FOR{edge communication round $e$ = 1,...,$E$}
    \FOR{local device $n$ = 1,...,$N_k$}
        \STATE Generate mask $\bm{m}_{k,n}^{q,e}$.
        \STATE Generate $\bm{w}_{k,n}^{q,e,0} = \bm{w}_{k,n}^{q,e} \odot\bm{m}_{k,n}^{q,e}$.
        \FOR{iteration $t$ = 1,2,...,$T$}
            \STATE Update $\bm{w}_{k,n}^{q,e,t+1}$ as (8).
        \ENDFOR
    \ENDFOR
    \FOR {parameter $j$ in local models}
        \STATE Find $\mathcal{N}_{k,q,e}^{j} = \{n: m_{k,n}^{q,e,j} = 1\}$.
        \STATE Update $\hat{w}_{k,q}^{e,T,j}$ as (12).
    \ENDFOR
\ENDFOR
\STATE Update $\bm{w}_{G}^{q+1}$ as (13).
\ENDFOR
\end{algorithmic}
\end{algorithm}

\subsection{Computation and Communication Latency}
Synchronous training is deployed in HFL and we mainly consider computation and uplink transmission latency. Note that the computational complexity of importance calculation is very low as compared with the local forward and back propagation during model training. Therefore, the computational complexity of importance calculation is ignored, and the latency for local computation and uplink transmission is written as 
\begin{align}
    T_{n,k,e} &= T_{n,e}^{\text{cmp}} + T_{n,k,e}^{\text{up}}\\
    &= \frac{TC_nW_{\rho_{n,e}}}{f_n} + \frac{\hat{q}W_{\rho_{n,e}}}{R_{n,k,e}^{\text{up}}}.
\end{align}
As a result, the latency of the $k$th edge server in the $e$th edge communication round is expressed as
\begin{equation}
    T_{k,e} = \max_{n\in\mathcal{S}_k}\{T_{n,k,e}\}.
\end{equation}
From (16), we observe that the bottleneck of the computation and communication latency is affected by the last device that finishes all local iterations and uplink transmission after local model updating.

\section{Convergence Analysis and Problem Formulation}
In this section, the convergence analysis of model pruning in HFL is first analyzed. Then, an optimization problem is formulated to minimize the upper bound of the convergence analysis.

\subsection{Convergence Analysis}
Since the neural network is non-convex in general, the average $l_2$-norm of gradients is deployed to evaluate the convergence performance \cite{Ghadimi, SShi}. The following assumptions are employed in hierarchical federated pruning convergence analysis.

$\textbf{Assumption 1.}$ (Smoothness) Cost functions $F_1,...,F_N$ are all $L-$ smooth:
\begin{equation}
    \|\nabla F_{n}(\bm{w}_1) - \nabla F_{n}(\bm{w}_2)\| \leq L\|\bm{w}_1 - \bm{w}_2\|,
\end{equation}
where $L$ is a positive constant.

$\textbf{Assumption 2.}$ (Pruning-induced Noise) Different from the other convergence analysis of HFL in \cite{HFEL_quantization,HFEL_SGD,Demystifying,HFEL_mobility}, we consider the effect of pruning-induced noise. According to \cite{Stich}, the model error of the $n$th device under the pruning ratio $\rho_{n,e}$ is bounded by 
\begin{equation}
    \mathbb{E}\|\bm{w}_{k,n}^{q,e} - \bm{w}_{k,n}^{q,e}\odot\bm{m}_{k,n}^{q,e}\|^2\leq\rho_{n,e}D^2,
\end{equation}
where $D$ is a positive constant.

$\textbf{Assumption 3.}$ (Bounded Gradient) The second moments of stochastic gradients is bounded \cite{Bartlett,Salimans}, which is denoted as
\begin{equation}
    \mathbb{E}\|\nabla F_{n}(\bm{w}_{k,n}^{q,e,t}, \xi_{k,n}^{q,e,t})\|^2\leq \phi^2.
\end{equation}
In (19), $\phi$ is a positive constant, and $\xi_{k,n}^{q,e,t}$ are mini-batch data samples for any $k,n,q,e,t$.

$\textbf{Assumption 4.}$ (Gradient Noise for IID data) Under IID data distribution, we assume that
\begin{equation}
    \mathbb{E}[\nabla F_{n}(\bm{w}_{k,n}^{q,e,t}, \xi_{k,n}^{q,e,t})] = \nabla F_{n}(\bm{w}_{k,n}^{q,e,t}),
\end{equation}
and
\begin{equation}
    \mathbb{E}\|\nabla F_{n}(\bm{w}_{k,n}^{q,e,t}, \xi_{k,n}^{q,e,t}) - \nabla F_{n}(\bm{w}_{k,n}^{q,e,t})\|^2\leq \hat{\sigma}^2.
\end{equation}
In (21), $\hat{\sigma} > 0$ is a constant.

\textbf{Theorem 1:} With the above assumptions, HFL with pruning converges to a small neighborhood of a stationary point of standard FL as follows:
\begin{align}
    \frac{1}{QW}\sum_{q=1}^{Q}\sum_{j=1}^{W}\mathbb{E}\|\nabla F^j(\bm{w}_{q})\|^2\leq \frac{2\mathbb{E}[F(\bm{w}^{0}) - F(\bm{w}^{*})]}{QW\eta ET} + H_1  + H_2\sum_{e=1}^{E}\sum_{n=1}^{N} \rho_{n,e},
\end{align}
where $H_1$ and $H_2$ are expressed as
\begin{align}
    H_1 = 3L\eta TEW\phi^2 +\frac{\phi^2N\eta^2T^2L^3 + 3LW\eta ETN\hat{\sigma}^2 + 3WEL^3T^3\eta^3\phi^2N}{\Gamma^{*}},
\end{align}
and
\begin{equation}
    H_2 = \frac{2EL^2 + 6W\eta L^3D^2T} {\Gamma^{*}}.
\end{equation}
In (22), (23), and (24), $Q$ is the number of global communication rounds, $E$ is the number of edge communication rounds, $T$ is the number of iterations in each device, $W$ is the total number of model weights, and $\Gamma^{*}$ is the minimum occurrence of the parameter in local models of all rounds.

$\mathit{Proof:}$ Please refer to Appendix A.

\subsection{Problem Formulation}
Based on the aforementioned system model and convergence analysis, we consider an optimization problem with respect to minimizing the upper bound in (22). The optimization problem is formulated as follows:
\begin{align}
    \min_{{b}_{n,e}, \rho_{n,e}}~~&H_2\sum_{e=1}^{E}\sum_{n=1}^{N} \rho_{n,e},\\
    s.t.~~&T_{n,k,e}\leq T_{\text{th}},\\
    ~~&\sum_{n=1}^{N}{b}_{n,e}\leq 1,\\
    ~~&0\leq {b}_{n,e}\leq 1,\\
    ~~&\rho_{n,e}\in [0, 1].
\end{align}
where $T_\text{th}$ in (26) represents the computation and communication latency constraint, (27) and (28) represent the wireless resource constraints, namely, the bandwidth fraction ${b}_{n,e}$ allocated to the $n$th device in the $e$th edge communication round cannot larger than the total bandwidth $B$, and (29) represents the pruning ratio constraint, which should be carefully selected to avoid the learning accuracy decreasing sharply.

Minimizing the global loss function requires an explicit form about how to select the pruning ratio based on the latency and wireless resource constraints. Since it is almost impossible to know the training performance exactly before the model has been trained, we turn to find an upper bound of $l_2$-norm of gradients and minimize it for the global loss minimization \cite{10089235}.  Obviously, the optimization problem in (25) is a mixed integer non-linear programming (MINLP) problem, which is non-convex and impractical to directly obtain optimal solutions. Therefore, we decompose the original problem into several sub-problems to obtain sub-optimal solutions.

\section{Optimal Pruning Ratio and Wireless Resource Allocation}
In this section, we decompose the optimization in (25) into two sub-problems with the aim to derive the optimal solutions of pruning ratio and wireless resource allocation.

\subsection{Optimal Pruning Ratio}
Based on (26), the transmission and computation latency of the $n$th mobile device should satisfy the latency threshold, which is denoted as
\begin{equation}
\frac{TC_nW_{\rho_{n,e}}}{f_n} + \frac{\hat{q}W_{\rho_{n,e}}}{R_{n,k,e}^{\text{up}}}\leq T_{\text{th}}.
\end{equation}

\textbf{Theorem 2:} The pruning ratio of the $n$th device associated with the $k$th edge server should satisfy
\begin{equation}
    \rho_{n,e}^{\star} \geq \left(1 - \frac{T_{\text{th}} - T_{n,e}^{\text{cmp-conv}} - T_{n,k,e}^{\text{com-conv}}}{T_{n,e}^{\text{cmp-fully}} + T_{n,k,e}^{\text{com-fully}}}\right)^{+},
\end{equation}
where $T_{n,e}^{\text{cmp-conv}}$ and $T_{n,k,e}^{\text{com-conv}}$ are computation and uplink transmission latency of the convolutional layer, respectively. In (31), $T_{n,e}^{\text{cmp-fully}}$ and $T_{n,k,e}^{\text{com-fully}}$ are computation and uplink transmission latency of the fully-connected layer, respectively. Also, in (31), $(z)^{+} = \max(z, 0)^{+}$.

$\mathit{Proof:}$ Please refer to Appendix B.

\textit{Remark 1:} From Theorem 2 and (59) in the Appendix B, the optimal pruning ratio is jointly determined by the computation capability and uplink transmission rate of the local device. For the local device with a high computation capability, the learning model can be pruned with a small pruning ratio to improve the convergence rate. Also, the pruning ratio decreases with a high uplink transmission rate. When more bandwidth is allocated to the local device, a small pruning ratio can be adopted.

\subsection{Optimal Wireless Resource Allocation}
According to the optimal pruning ratio in (31), the optimization problem in (25) is rewritten as
\begin{equation}
    \min~~H_2\sum_{e=1}^{E}\sum_{n=1}^{N}\left(1 - \frac{T_{\text{th}} - T_{n,e}^{\text{cmp-conv}} - T_{n,k,e}^{\text{com-conv}}}{T_{n,e}^{\text{cmp-fully}} + T_{n,k,e}^{\text{com-fully}}}\right),
\end{equation}
subject to (27) and (28). When $T_{n,k,e}^{\text{com-conv}} = W_{n,\text{conv}}\hat{q}/R_{n,k,e}^{\text{up}}$ and $T_{n,k,e}^{\text{com-fully}} = W_{n,\text{com-fully}}\hat{q}/R_{n,k,e}^{\text{up}}$, (32) is further derived as
\begin{equation}
    \min~~H_2\sum_{e=1}^{E}\sum_{n=1}^{N}\left(1 - \frac{R_{n,k,e}^{\text{up}}(T_{\text{th}} - T_{n,e}^{\text{cmp-conv}}) - \hat{q}W_{n,\text{conv}}}{R_{n,k,e}^{\text{up}}T_{n,e}^{\text{cmp-fully}} + \hat{q}W_{n,\text{fully}}}\right).
\end{equation}
The optimal solution of wireless resource allocation is obtained by solving the optimization problem in (33). First, according to the following Lemma 1, we prove that the optimization problem in (33) is convex.

$\textbf{Lemma~1:}$ The optimization problem in (33) is convex.

$\mathit{Proof:}$ Please refer to Appendix C.

According to the Lagrange multiplier method, the optimal wireless resource allocation is obtained in the following theorem.

\textbf{Theorem 3:} To obtain the optimal learning performance, the optimal bandwidth allocated to the $n$th deivce should satisfy
\begin{equation}
    {b}_{n,e}^{\star} = \frac{\sqrt{\frac{(V_1V_4 + V_2V_3)B\log_{2}\left(1 + \frac{g_{n,k}^{e}p_n}{\sigma^2}\right)}{\lambda^{\star}}} - V_4}{BV_3\log_{2}\left(1 + \frac{g_{n,k}^{e}p_n}{\sigma^2}\right)},
\end{equation}
where $V_1 = T_{\text{th}} - T_{n,e}^{\text{cmp-conv}}$, $V_2 = \hat{q}W_{n,\text{conv}}$, $V_3 = T_{n,e}^{\text{cmp-fully}}$, $V_4 = \hat{q}W_{n,\text{fully}}$, and $\lambda^{\star}$ is the optimal Lagrange multiplier.

$\mathit{Proof:}$ Please refer to Appendix D.

Based on Theorem 2 and 3, the optimal pruning ratio is written as
\begin{equation}
    \rho_{n,e}^{\star}\!\! = \!\!1 - \frac{{b}_{n,e}^{\star}(T_{\text{th}} - T_{n,e}^{\text{cmp-conv}})B\log_{2}\!\!\left(\!1 \!\!+\!\! \frac{g_{n,k}^{e}p_n}{\sigma^2}\right) \!\!-\!\! \hat{q}W_{n,\text{conv}}}{{b}_{n,e}^{\star}T_{n,e}^{\text{cmp-fully}}B\log_{2}\left(1 + \frac{g_{n,k}^{e}p_n}{\sigma^2}\right) + \hat{q}W_{n,\text{fully}}}.
\end{equation}

\textit{Remark 2:} From Theorem 3, the wireless resource allocation decreases with better channel condition. More wireless resource is allocated to the local devices with bad channel condition to guarantee transmission latency. In addition, more wireless resource is allocated to the local devices with high computation capability, which can decrease the communication latency and improve the convergence rate.

\begin{table}
\centering
\caption{Simulation Parameters of HFL with Adaptive Model Pruning and Wireless Resource Allocation}
\begin{tabular}[c]{c|c|c|c}
\hline
\hline Transmission power of device & $28~\text{dBm}$ & Bandwidth & 20MHz  \\
\hline CPU frequency of device & $3~\text{GHz}$ & Learning rate & 0.001 \\
\hline AWGN noise power & -110~\text{dBm} & Batchsize & 128\\
\hline Quantization bit & 64 & Number of global communication rounds & 10\\
\hline Number of edge servers & 5 & Number of edge communication rounds & 5\\ 
\hline Number of devices per edge server & 5 & Number of local epochs & 2 \\
\hline
\hline
\end{tabular}\vspace{-10mm}
\end{table}

\section{Simulation Results}
In this section, we examine the effectiveness of our proposed HFL with model pruning via simulation. In the simulation, we consider a scenario with five edge servers and each edge server has five devices participating in model training. We deploy a common CNN model for image classification over the datasets MNIST and Fashion MNIST, which contain 50000 training samples and 10000 testing samples, respectively. The training data are shuffled to guarantee the local data are IID. The input size of CNN is $1 \times 28 \times 28$, and the sizes of the first and second convolutional layers are $32 \times 28 \times 28$ and $64 \times 14 \times 14$, respectively. The sizes of the first and second max-pooling layers are $32 \times 14 \times 14$ and $64 \times 7 \times 7$, respectively. The sizes of the first and second fully-connected layers are 3136 and 8, respectively. The size of the output layer is 10. The devices exchange learning models with edge servers over the wireless channel. The main simulation parameters are listed in Table I.\vspace{-10mm}

\begin{figure*}[ht]
	\centering
	\subfloat[]{\includegraphics[width=2.1in]{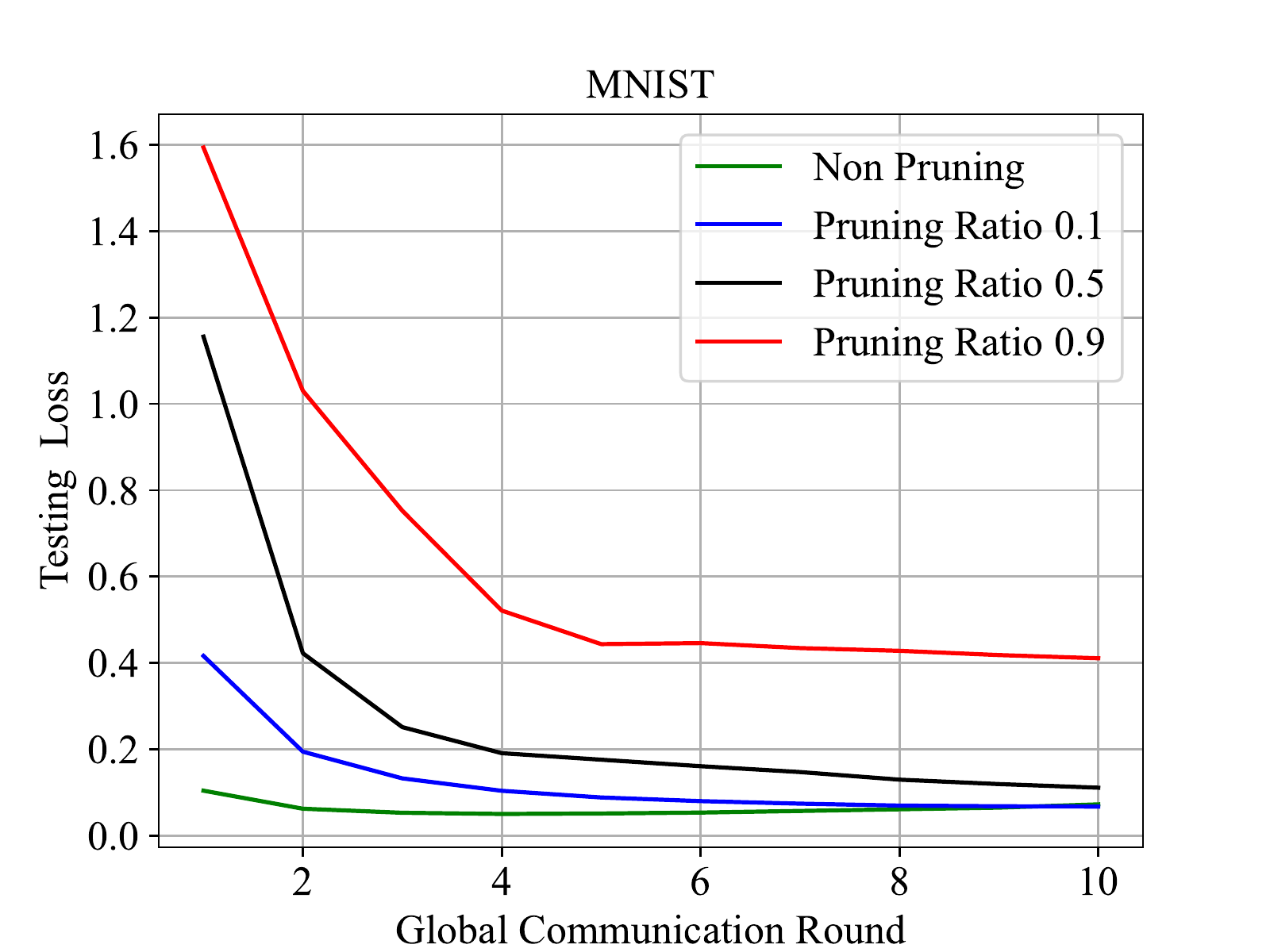}}\label{fig_first_case}
	\hfil
	\subfloat[]{\includegraphics[width=2.1in]{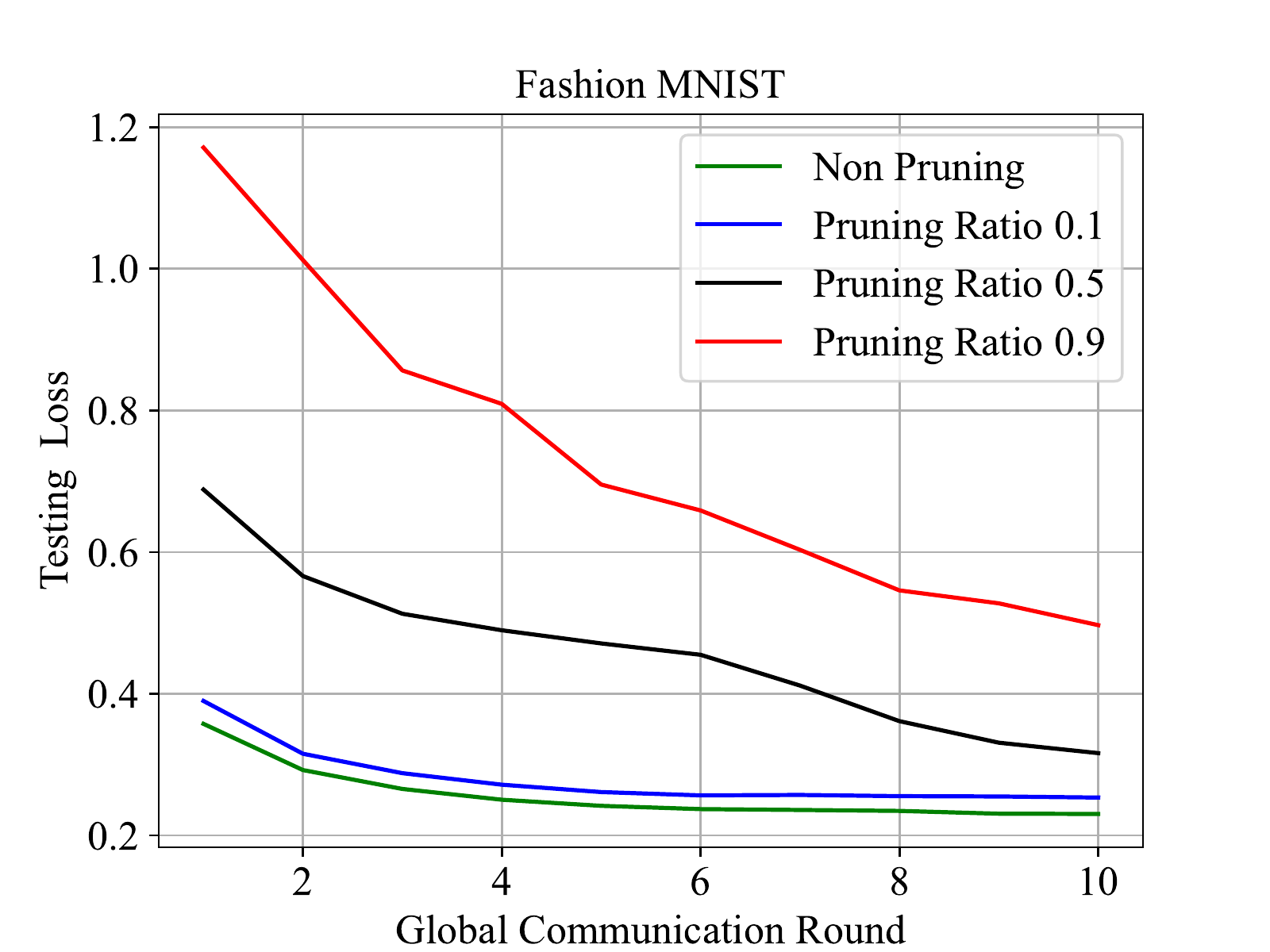}}\label{fig_second_case}
    \hfil
	\subfloat[]{\includegraphics[width=2.1in]{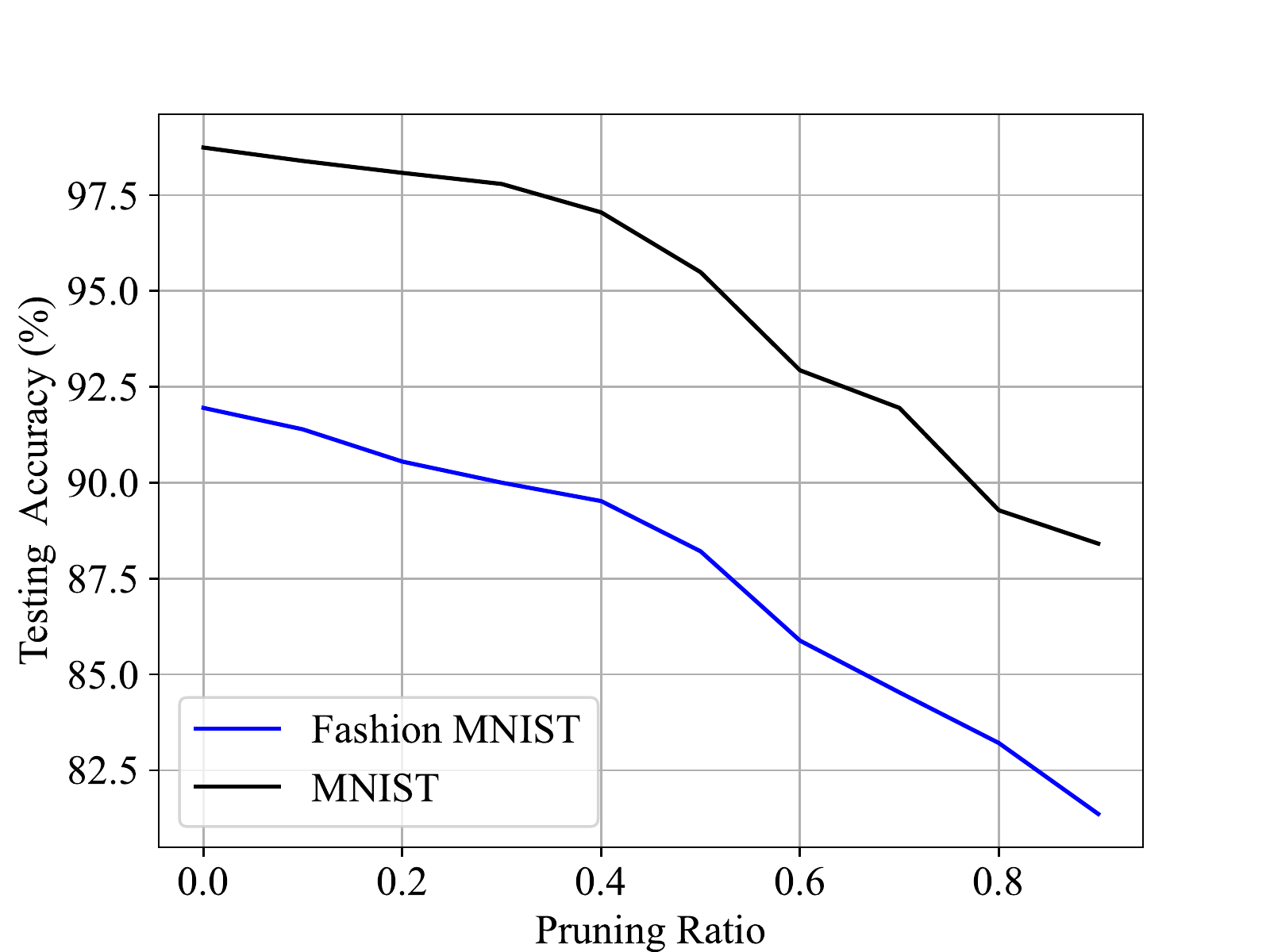}}\label{fig_third_case}
	\caption{(a) Testing Loss of importance-based model pruning of HFL with different pruning ratios on MNIST. (b) Testing Loss of importance-based model pruning of HFL with different pruning ratios on Fashion MNIST. (c) Testing accuracy of importance-based model pruning of HFL with different pruning ratios on MNIST and Fashion MNIST.}\vspace{-10mm}
	\label{basic_modules}
\end{figure*}

\subsection{HFL with Importance-based Pruning}
Fig. 2 (a) and Fig. 2 (b) plot the testing loss of importance-based model pruning of HFL with different pruning ratios on datasets MNIST and Fashion MNIST, respectively. Fig. 2 (c) plots the testing accuracy of importance-based model pruning of HFL under different pruning ratios on datasets MNIST and Fashion MNIST. It is observed that the testing loss increases, and convergence rate and testing accuracy decrease with increasing value of the pruning ratio. It is because a larger pruning ratio indicates that more weights may be pruned, which results in a high model aggregation error, and more iterations are required to train learning models.\vspace{-10mm}

\begin{figure*}[ht]
	\centering
	\subfloat[]{\includegraphics[width=2.1in]{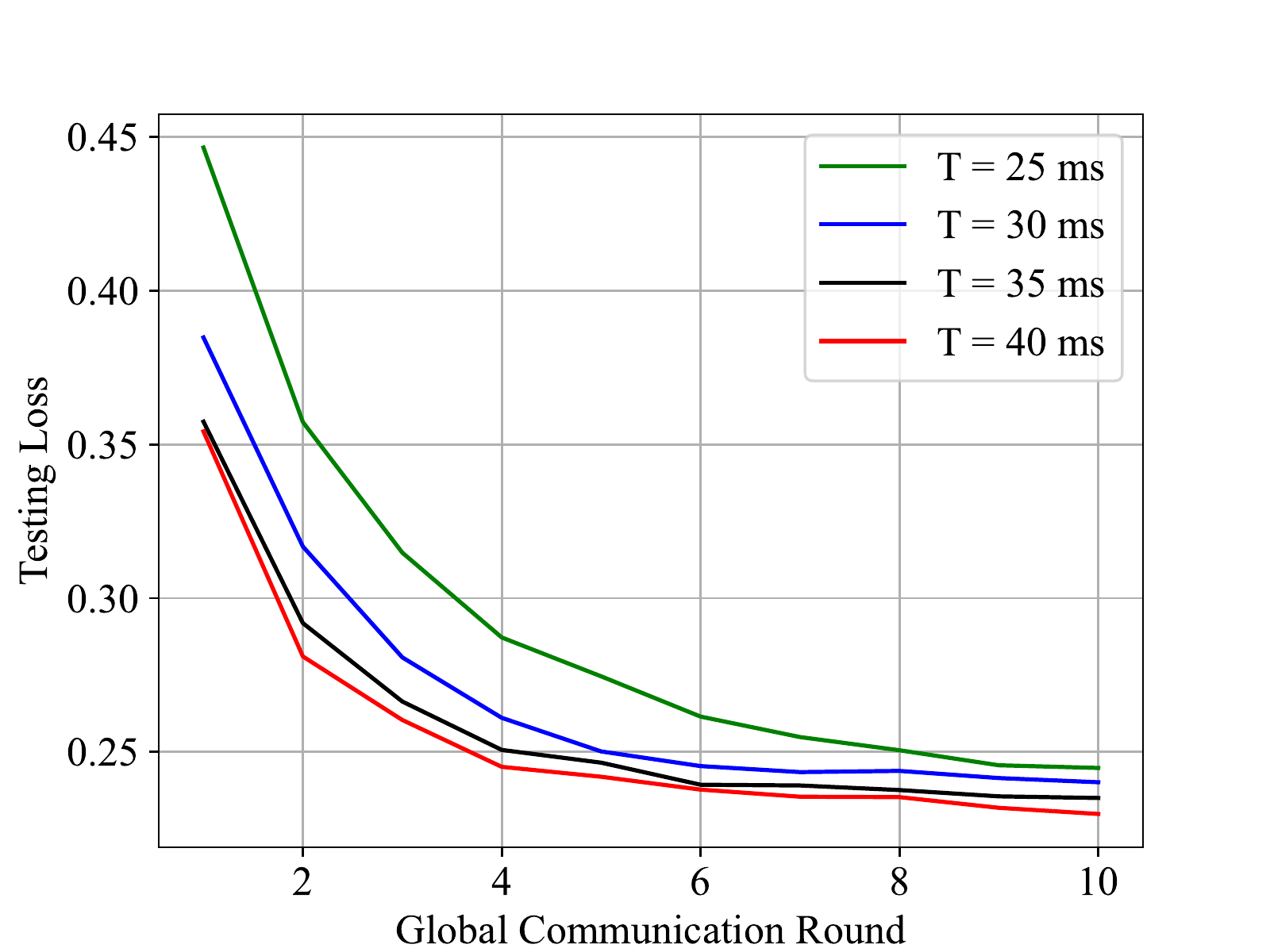}}\label{fig_first_case}
	\hfil
	\subfloat[]{\includegraphics[width=2.1in]{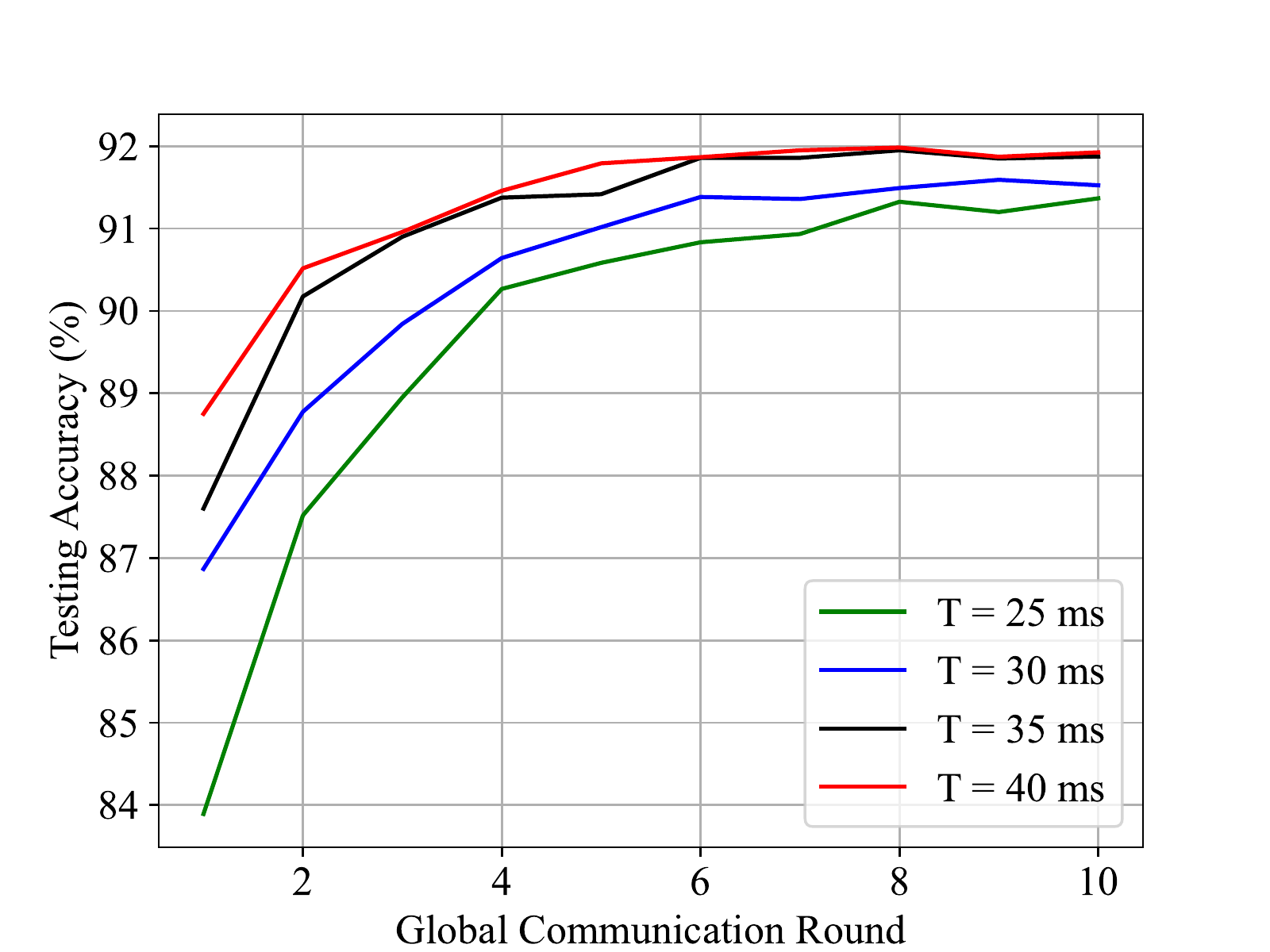}}\label{fig_second_case}
    \hfil
	\subfloat[]{\includegraphics[width=2.1in]{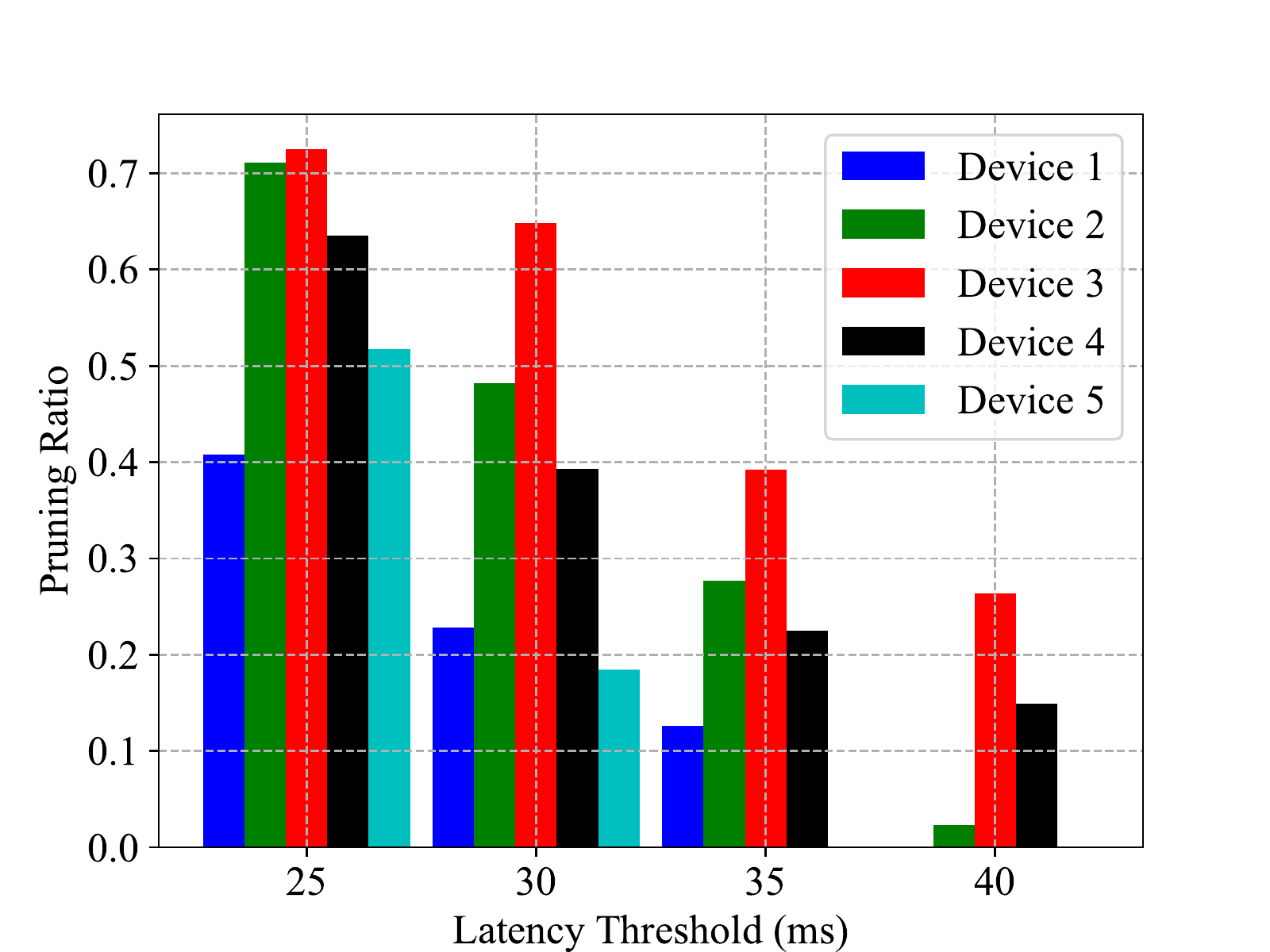}}\label{fig_third_case}
	\caption{(a) Testing loss of adaptive HFL model pruning with different latency thresholds on Fashion MNIST. (b) Testing accuracy of adaptive HFL model pruning with different latency thresholds on Fashion MNIST. (c) Pruning ratio required to achieve a given latency threshold on Fashion MNIST.}\vspace{-10mm}
	\label{basic_modules}
\end{figure*}

\subsection{HFL with Adaptive Model Pruning in Wireless Networks}
In this section, the effect of latency constraint in adaptive HFL model pruning and joint design of adaptive model pruning and wireless resource allocation over the dataset Fashion MNIST are simulated.

\subsubsection{Effect of Latency Constraint}
Fig. 3 (a) and Fig. 3 (b) present testing loss and accuracy of adaptive HFL model pruning with different latency constraints on Fashion MNIST, respectively. Fig. 3 (c) plots pruning ratio required for each device in one of edge servers to achieve a given latency threshold on Fashion MNIST. We consider four latency constraints, which are $25ms$, $30ms$, $35ms$, and $40ms$. From the figure, we observe that when the latency constraint increases, the testing loss decreases and the testing accuracy increases. Also, a small number of iterations is required to achieve convergence with a high latency constraint. In addition, the pruning ratio decreases with high latency thresholds. It is because with a large latency constraint, a small pruning ratio is selected by devices. On the contrary, for the device with a small latency constraint, a large pruning ratio is selected to satisfy the latency requirement while sacrificing the learning performance and more iterations are required to achieve convergence. In the following simulation, we assume that the latency constraint is $30ms$.

\subsubsection{Adaptive Model Pruning and Wireless Resource Allocation}
To demonstrate the joint design of adaptive model pruning and wireless resource allocation, we compare the proposed adaptive model pruning with other two baseline schemes. These three schemes are described as follows.
\begin{itemize}
    \item \textbf{Optimal Pruning}: Both the pruning ratio and wireless resource allocation are optimized according to Section V.
    \item \textbf{Equal Resource Pruning}: The pruning ratio is optimized. However, the bandwidth is equally allocated to all devices.
    \item \textbf{No Pruning}: The bandwidth is equally allocated to all devices and model pruning is not deployed.
\end{itemize}

\begin{table}
\centering
\caption{Per-round computation and communication latency (ms)}
\begin{tabular}[c]{c|c|c|c}
\hline
\hline Scheme & Optimal Pruning & Equal Resource Pruning & No Pruning \\
\hline HFL & 30 & 48 $\pm$ 0.21 & 52 $\pm$ 0.33  \\
\hline Non HFL & 350 & 590 $\pm$ 12.34 & 750 $\pm$ 15.44\\

\hline
\hline
\end{tabular}\vspace{-10mm}
\end{table}

Table II shows the per-round computation and communication latency of HFL and non-HFL schemes. It is observed that the computation and communication latency of Non-HFL is more than 10 times that of HFL. It is because in non-HFL networks, devices need to communication with the cloud server which is far away from them, which results in high transmission latency. Also, we can obtain that the computation and communication latency of optimal pruning is smaller than that of equal resource pruning and no pruning. This is because the optimal pruning ratio is selected based on the given wireless resource, which further decreases the computation and communication latency.  \vspace{-10mm}

\begin{figure*}[ht]
	\centering
	\subfloat[]{\includegraphics[width=2.1in]{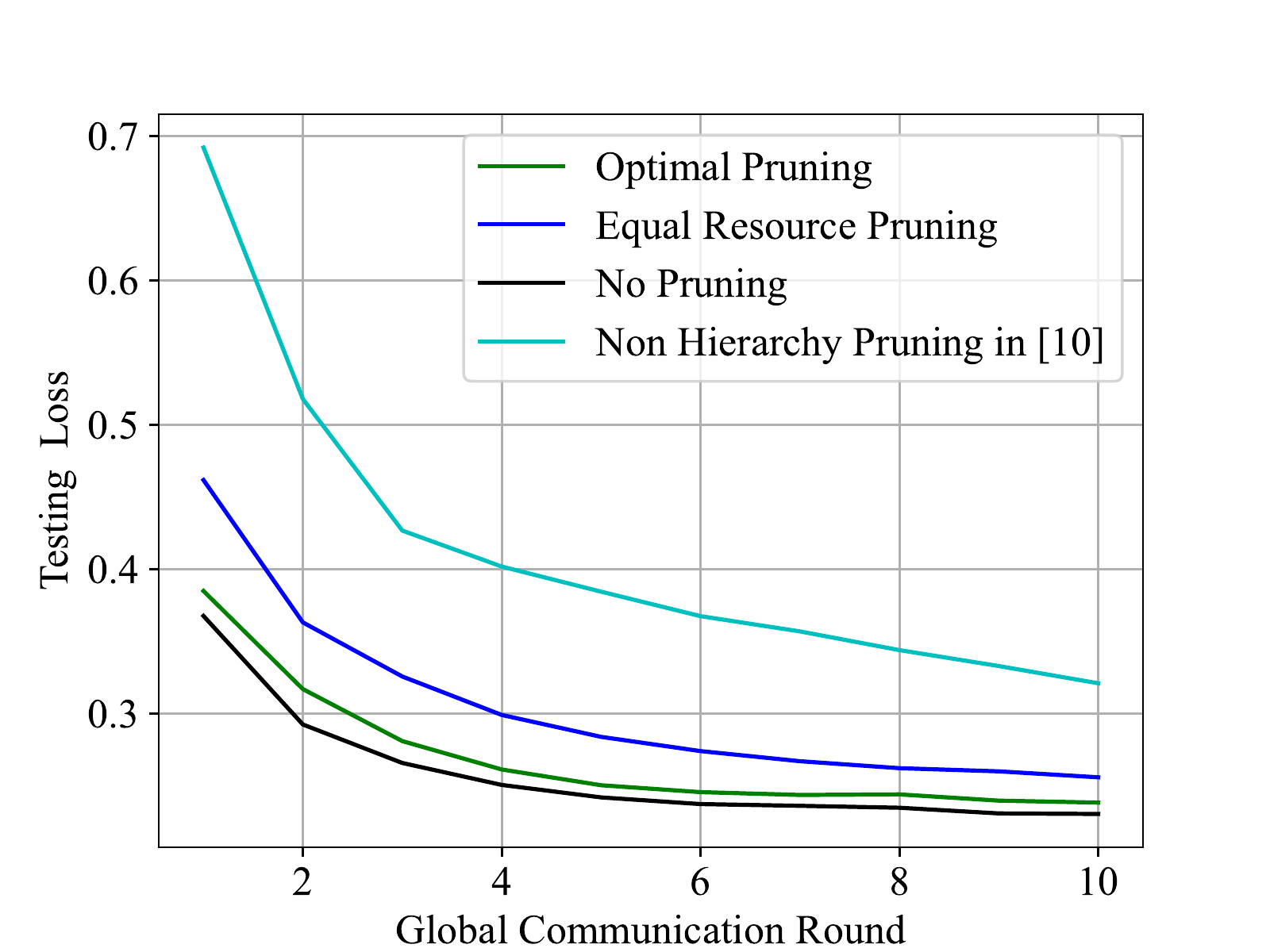}}\label{fig_first_case}
	\hfil
	\subfloat[]{\includegraphics[width=2.1in]{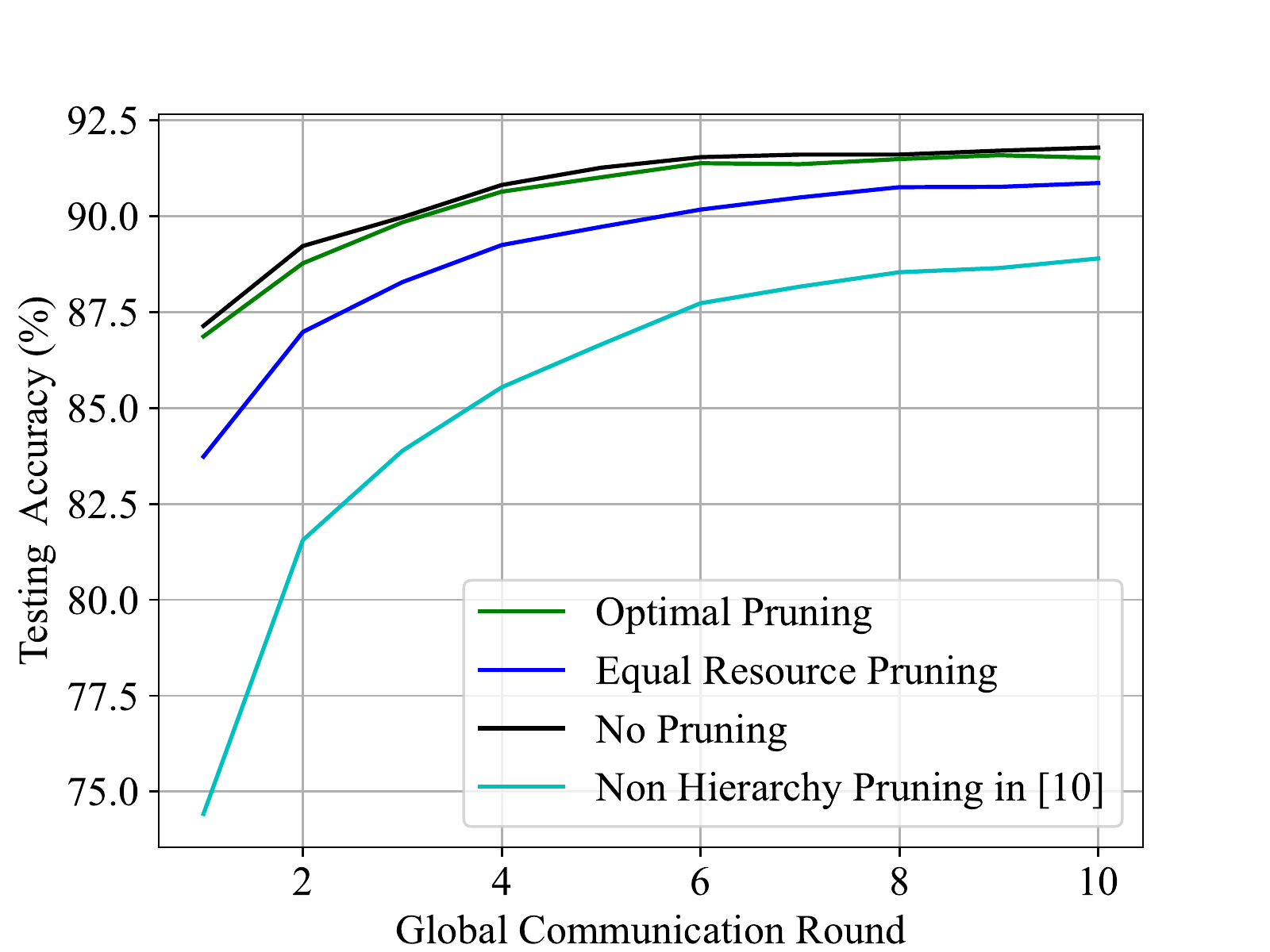}}\label{fig_second_case}
    \hfil
	\subfloat[]{\includegraphics[width=2.1in]{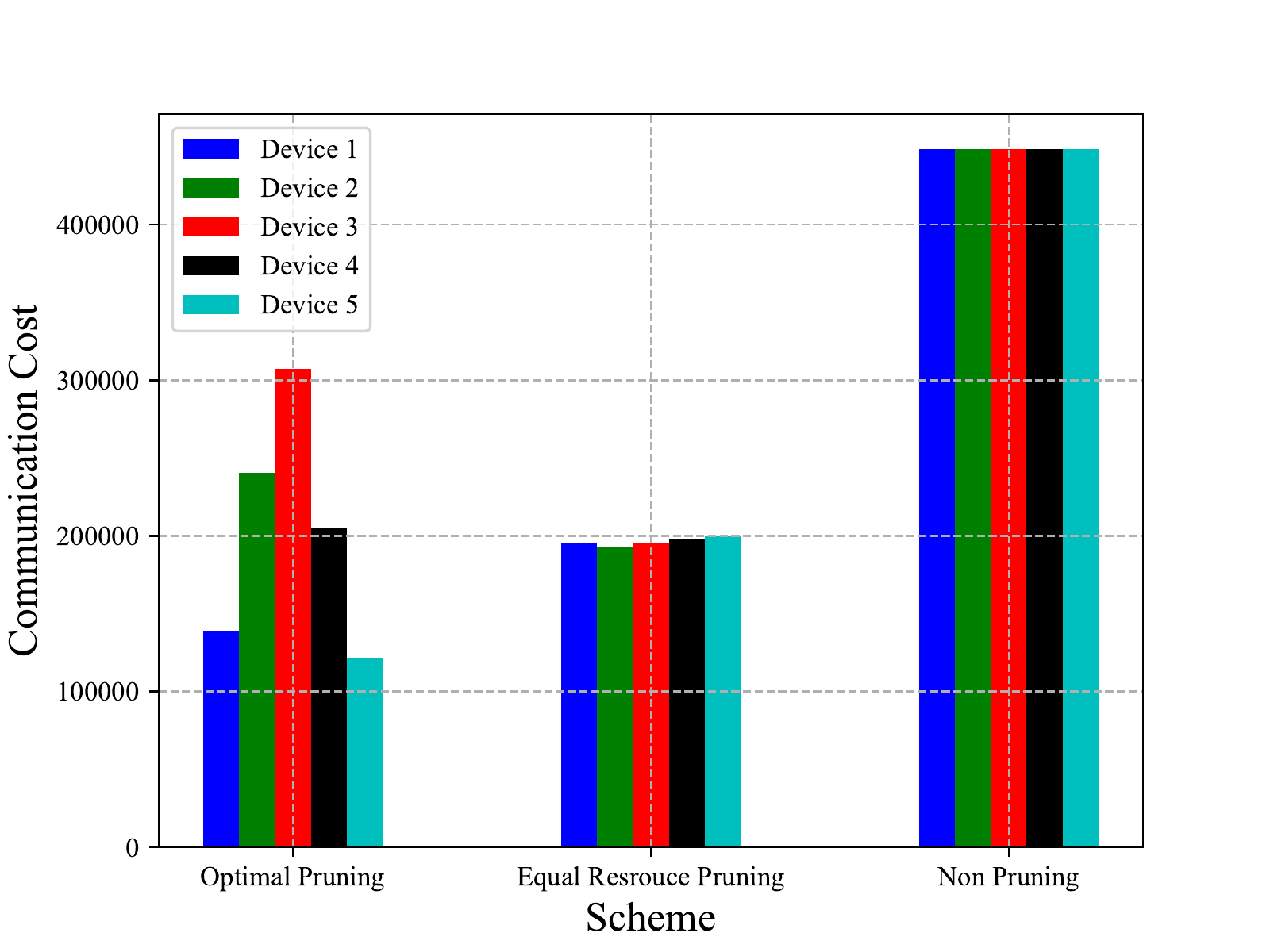}}\label{fig_third_case}
	\caption{(a) Testing loss of joint design of adaptive model pruning and wireless resource allocation on Fashion MNIST. (b) Testing accuracy of joint design of adaptive model pruning and wireless resource allocation on Fashion MNIST. (c) Communication costs on different schemes.}\vspace{-8mm}
	\label{basic_modules}
\end{figure*}

Fig. 4 (a) and Fig. 4 (b) plot the testing loss and accuracy of joint design of adaptive model pruning and wireless resource allocation on Fashion MNIST, respectively. Fig. 4 (c) plots communication costs on different schemes. The communication cost means the number of model weights needs to be uploaded. It is observed that optimal pruning has the ability to adapt to the wireless resource and the communication cost is much smaller than that of the no pruning scheme. These figures show that the performance of testing loss and accuracy of the proposed adaptive model pruning and wireless resource allocation is close to the no pruning scheme. However, the latency of the proposed algorithm is about $50\%$ less than that of the no pruning scheme. It is because the proposed adaptive pruning scheme has the ability to dynamically remove the unimportant weights according to the wireless channel, which further reduces the latency for both local model updating and uplink transmission, especially when the model size is large. In addition, from Fig. 4 (a) and Fig. 4 (b), we can obtain that the learning accuracy of HFL with the optimal model pruning is much better than that of model pruning in non-hierarchy networks \cite{9598845}. This is because, in HFL, the participating devices are able to provide massive datasets for model updating, which further improves the learning performance. \vspace{-10mm}

\begin{figure*}[ht]
	\centering
	\subfloat[]{\includegraphics[width=3.0in]{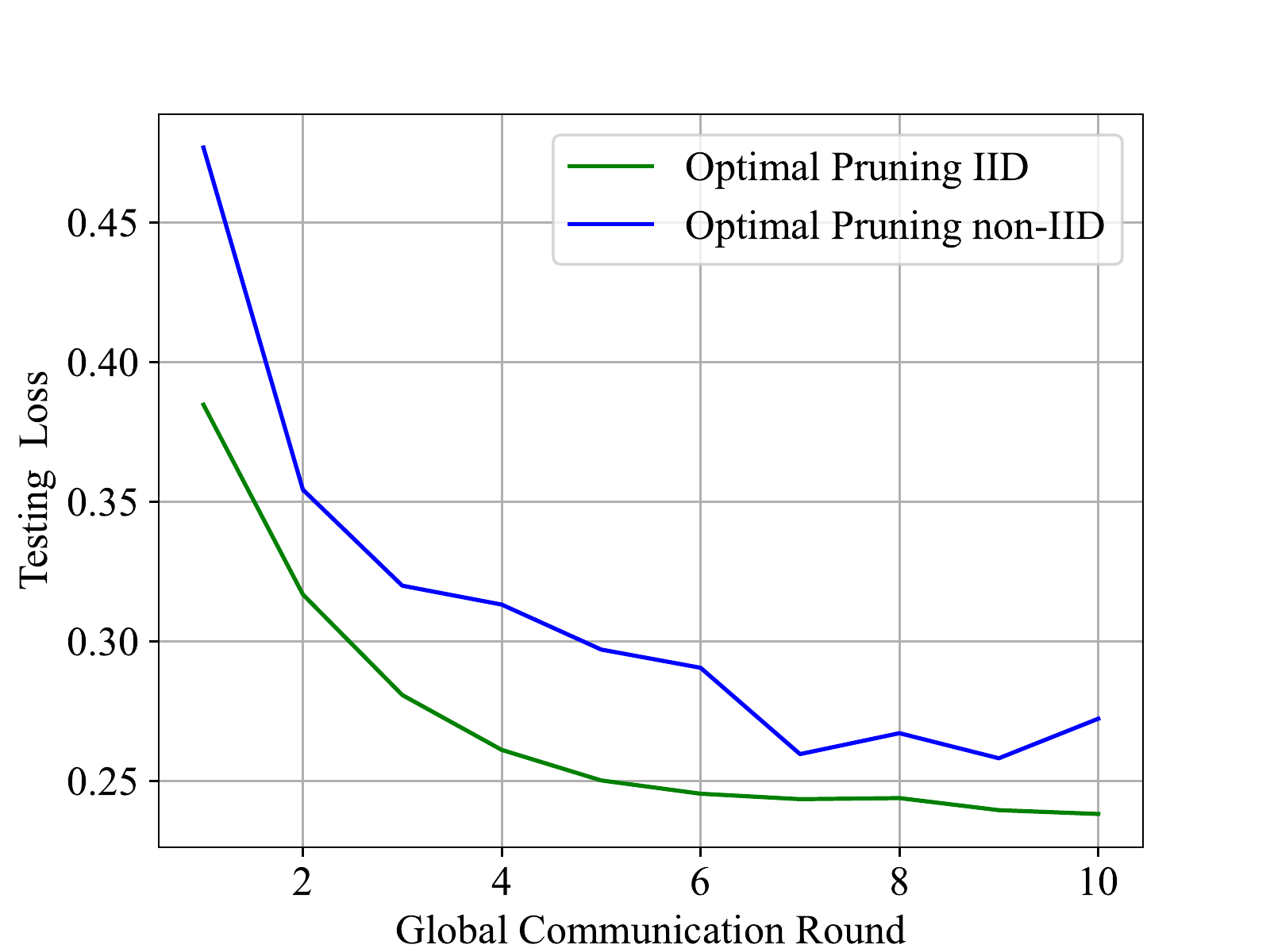}}\label{fig_first_case}
	\hfil
	\subfloat[]{\includegraphics[width=3.0in]{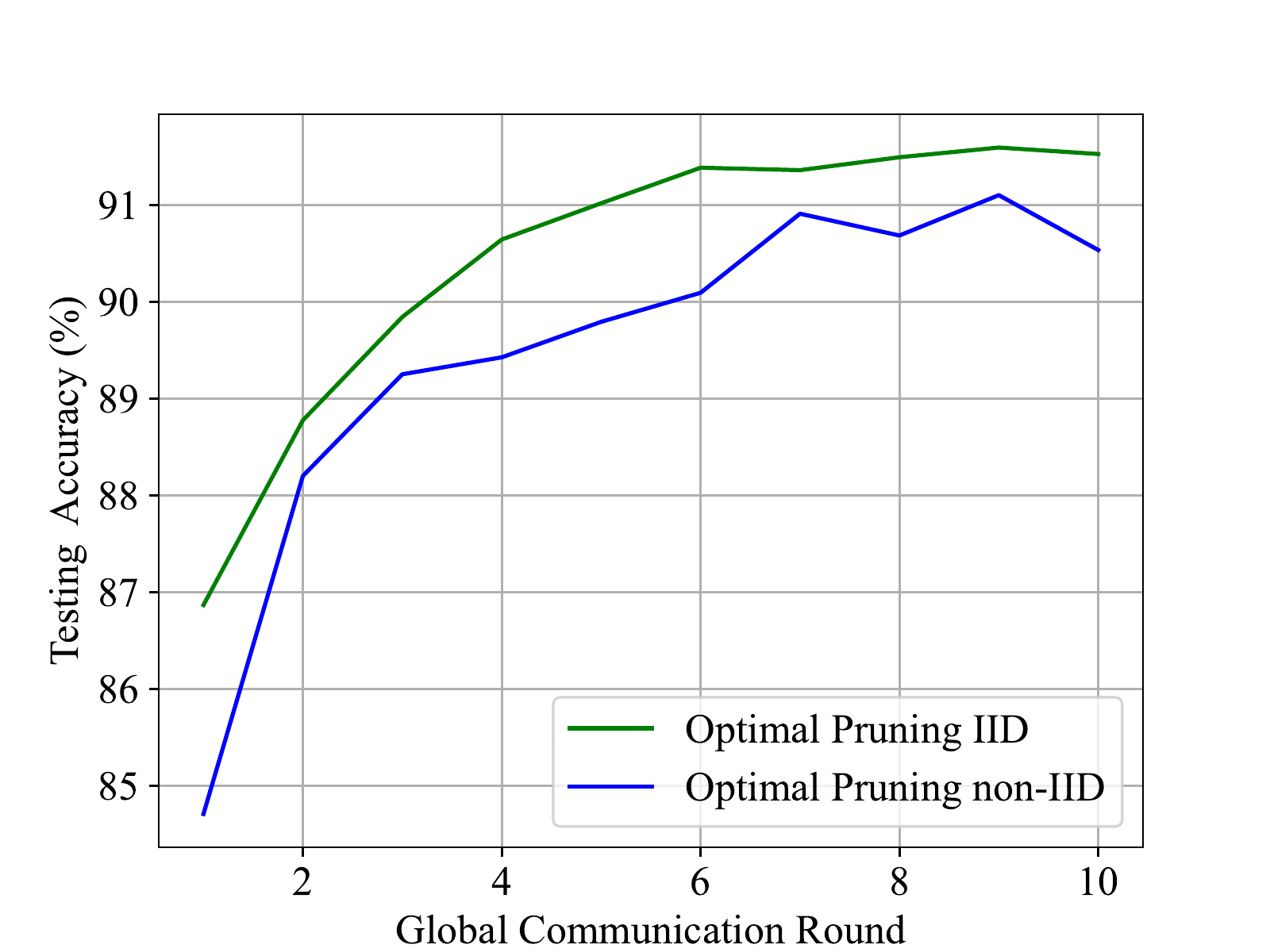}}\label{fig_second_case}
	\caption{(a) Testing loss of joint design of adaptive model pruning and wireless resource allocation on IID or non-IID Fashion MNIST. (b) Testing accuracy of joint design of adaptive model pruning and wireless resource allocation on IID or non-IID Fashion MNIST.}\vspace{-6mm}
	\label{basic_modules}
\end{figure*}

Fig. 5 (a) and Fig. 5 (b) plot the testing loss and accuracy of joint design of adaptive model pruning and wireless resource allocation on IID or non-IID Fashion MNIST. It is observed that the testing accuracy and loss of IID data are better than that of non-IID data. It is because in non-IID data, the data class distribution at each device is skewed, which means some data classes are too scarce or even missing. Also, servers need more communication rounds to converge.

\begin{figure}[!h]
    \centering
    \includegraphics[width=3.5 in]{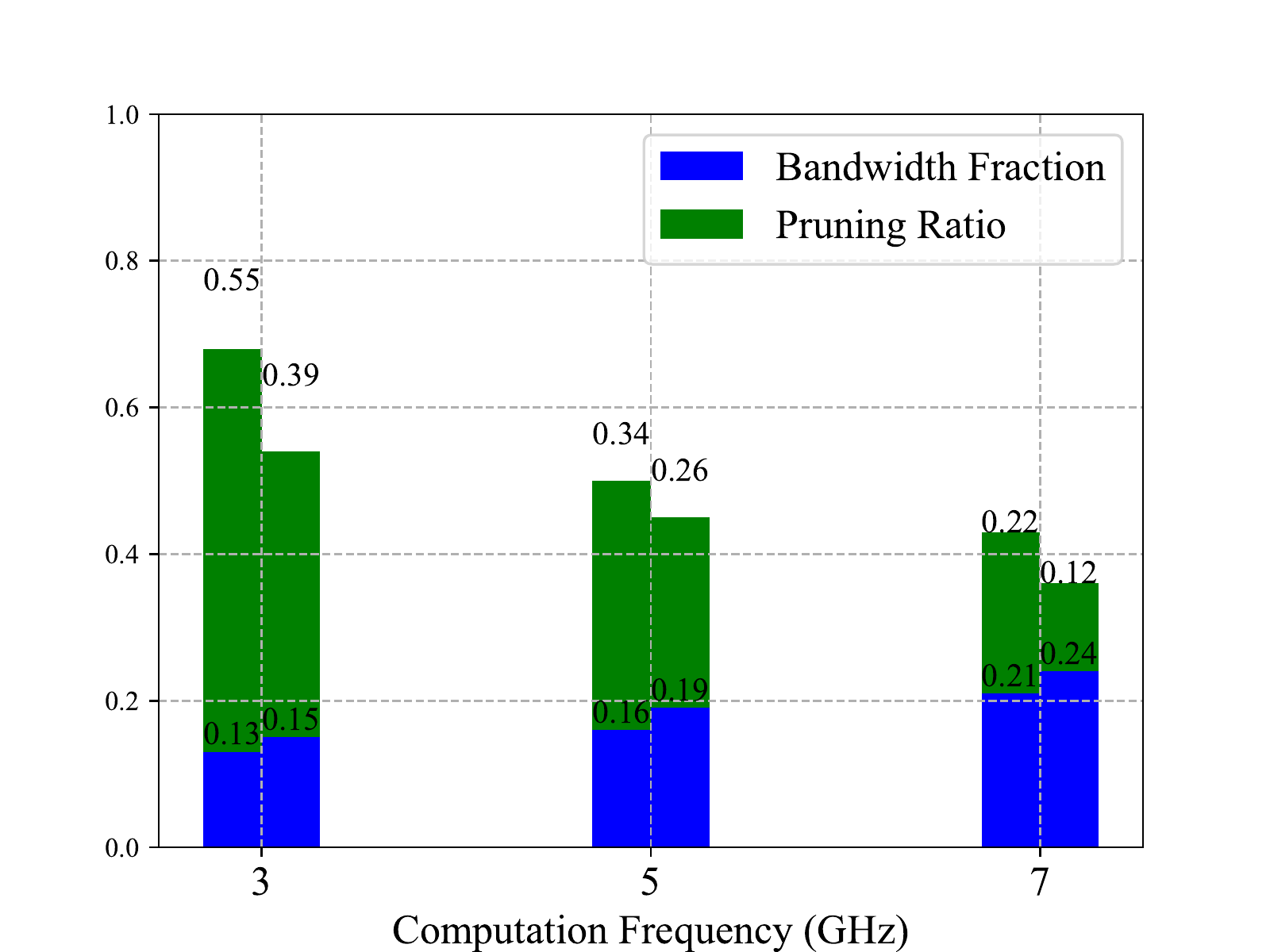}
    \caption{Relationship among pruning ratio, wireless resource allocation, and computation capability in the proposed optimal pruning scheme.}\vspace{-10mm}
    \label{basic_modules}
\end{figure}

Fig. 6 plots the relationship among pruning ratio, wireless resource allocation, and computation capability in the proposed optimal pruning scheme. It is observed that under the same computation capability, when more bandwidth is allocated to the local device, a smaller pruning ratio is adopted to guarantee a high convergence rate. Also, we can obtain that for the local device with a higher computation capability, more wireless resource is allocated to the local device, and a smaller pruning ratio is selected to guarantee the computation and communication latency and improve the convergence rate.

\section{Conclusions}
In this paper, an adaptive model pruning for HFL in wireless networks was developed to reduce the learning network scale. Specifically, the convergence analysis of an upper bound on the $l_2$-norm of gradients for HFL with model pruning was derived. Then, the pruning ratio and wireless resource allocation were jointly optimized under latency and bandwidth constraints by KKT conditions. Simulation results have shown that our proposed HFL with model pruning achieved similar learning accuracy compared to HFL without pruning and reduced about $50\%$ computation and communication latency.

\appendix
\section{Appendix }
\subsection{Appendix A - Proof of Theorem 1}
We now analyze the convergence of HFL with respect to the pruning ratio $\rho$ and pruning mask $\bm{m}$. Throughout the proof, we use the following inequalities frequently.

From Jensen's inequality, for any $\bm{z}_{m}\in\mathbb{R}^{d}, m\in\{1, 2, ..., M\}$, we have
\begin{equation}
    \left\|\frac{1}{M}\sum_{m=1}^{M}\bm{z}_{m}\right\|^2\leq \frac{1}{M}\sum_{m=1}^{M}\|\bm{z}_{m}\|^2,
\end{equation}
which directly gives
\begin{equation}
    \left\|\sum_{m=1}^{M}\bm{z}_{m}\right\|^2\leq M\sum_{m=1}^{M}\|\bm{z}_{m}\|^2.
\end{equation}
Peter-Paul inequality (also known as Young's inequality) gives
\begin{equation}
\langle\bm{z}_1,\bm{z}_2\rangle \leq \frac{1}{2}\|\bm{z}_1\|^2 + \frac{1}{2}\|\bm{z}_2\|^2,
\end{equation}

and for any constant $s>0$ and $\bm{z}_{1}, \bm{z}_{2}\in\mathbb{R}^{d}$, we have
\begin{equation}
    \|\bm{z}_{1} + \bm{z}_{2}\|^2\leq (1+s)\|\bm{z}_{1}\|^2 + \left(1 + \frac{1}{s}\right)\|\bm{z}_{2}\|^2.
\end{equation}

\textbf{Lemma 2:} Under Assumption 2 and 3, for any global and edge communication rounds $q$ and $e$, we obtain that
\begin{align}
    \sum_{e=1}^{E}\sum_{t=1}^{T}\sum_{n=1}^{N}\mathbb{E}\|\bm{w}_{k,n}^{q,e,t-1} - \bm{w}_{k,n}^{q,e}\|^2\leq \eta^2 \phi^2NET^3 + 2TD^2\sum_{e=1}^{E}\sum_{n=1}^{N}\rho_{n,e}.
\end{align}
$\mathit{Proof:}$ In (40), $\bm{w}_{k,n}^{q,e}$ is the received edge model of the $k$th edge server from the cloud server at the beginning of the $e$th edge communication round, and difference $(\bm{w}_{k,n}^{q,e,t-1} - \bm{w}_{k,n}^{q,e})$ consists of two parts, namely, variation because of local model training $(\bm{w}_{k,n}^{q,e,t-1} - \bm{w}_{k,n}^{q,e,0})$ and variation because of pruning $(\bm{w}_{k,n}^{q,e,0} - \bm{w}_{k,n}^{q,e})$. Therefore, (40) is rewritten as

\begin{align}
    &\sum_{e=1}^{E}\sum_{t=1}^{T}\sum_{n=1}^{N}\mathbb{E}\|\bm{w}_{k,n}^{q,e,t-1} - \bm{w}_{k,n}^{q,e}\|^2  = \sum_{e=1}^{E}\sum_{t=1}^{T}\sum_{n=1}^{N}\mathbb{E}\|(\bm{w}_{k,n}^{q,e,t-1} \!\!\!\!-\! \bm{w}_{k,n}^{q,e,0}) + (\bm{w}_{k,n}^{q,e,0}\!\!\!\! - \!\bm{w}_{k,n}^{q,e})\|^2 \nonumber\\
    \leq &\sum_{e=1}^{E}\sum_{t=1}^{T}\sum_{n=1}^{N}2\mathbb{E}\|\bm{w}_{k,n}^{q,e,t-1} - \bm{w}_{k,n}^{q,e,0}\|^2 +\sum_{e=1}^{E}\sum_{t=1}^{T}\sum_{n=1}^{N}2\mathbb{E}\|\bm{w}_{k,n}^{q,e,0} - \bm{w}_{k,n}^{q,e}\|^2.
\end{align}

In (41), $\bm{w}_{k,n}^{q,e,t-1}$ is updated from $\bm{w}_{k,n}^{q,e,0}$ by $t-1$ iterations on the $n$th device. Through the local gradient updating, we obtain that

\begin{align}
    &\sum_{e=1}^{E}\sum_{t=1}^{T}\sum_{n=1}^{N}2\mathbb{E}\|\bm{w}_{k,n}^{q,e,t-1} - \bm{w}_{k,n}^{q,e,0}\|^2 = \sum_{e=1}^{E}\sum_{t=1}^{T}\sum_{n=1}^{N}2\mathbb{E}\left\|\sum_{i=0}^{t-2}-\eta\nabla F_n(\bm{w}_{k,n}^{q,e,i}, \xi_{k,n}^{q,e,i})\odot\bm{m}_{k,n}^{q,e}\right\|^2\nonumber\\
    &\leq 2\eta^2\sum_{e=1}^{E}\sum_{t=1}^{T}\sum_{n=1}^{N}(t-1)\sum_{i=0}^{t-2}\mathbb{E}\|\nabla F_n(\bm{w}_{k,n}^{q,e,i}, \xi_{k,n}^{q,e,i})\odot\bm{m}_{k,n}^{q,e}\|^2\nonumber\\
    &\leq 2\eta^2 \phi^2NE\sum_{t=1}^{T}(t-1)^2
    =\eta^2 \phi^2NE\frac{2T^3 - 3T^2 + T}{3}\leq\eta^2 \phi^2NET^3,
\end{align}
where the third step in (42) is obtained from the bounded gradient in Assumption 3.

Then, $\bm{w}_{k,n}^{q,e,0} - \bm{w}_{k,n}^{q,e}$ in (41) is calculated as
\begin{align}
    &\sum_{e=1}^{E}\sum_{t=1}^{T}\sum_{n=1}^{N}2\mathbb{E}\|\bm{w}_{k,n}^{q,e,0} - \bm{w}_{k,n}^{q,e}\|^2 =\sum_{e=1}^{E}\sum_{t=1}^{T}\sum_{n=1}^{N}2\mathbb{E}\|\bm{w}_{k,n}^{q,e} \odot\bm{m}_{k,n}^{q,e} - \bm{w}_{k,n}^{q,e}\|^2 \nonumber\\
    &\leq 2\sum_{e=1}^{E}\sum_{t=1}^{T}\sum_{n=1}^{N}\rho_{n,e}D^2 = 2TD^2\sum_{e=1}^{E}\sum_{n=1}^{N}\rho_{n,e},
\end{align}
where the second step is obtained from pruning-induced noise in Assumption 2.
By plugging (42) and (43) into (41), we obtain the desired result, which ends the proof of Lemma 2.

\textbf{Lemma 3:} Under Assumptions 1-3, for any global and edge communication rounds $q$ and $e$, we obtain that
\begin{align}
    &\mathbb{E}\left\|\sum_{e=1}^{E}\frac{1}{\Gamma_{k}^{q,e,j}}\sum_{t=1}^{T}\sum_{n\in\mathcal{N}_{k}^{q,e,j}}[\nabla F_{n}^{j}(\bm{w}_{k,n}^{q,e,t-1}) - \nabla F_{n}^{j}(\bm{w}_{k,n}^{q,e})]\right\|^2 \nonumber\\
    &\leq \frac{\phi^2NE^2\eta^2L^2T^4 + 2ET^2L^2D^2\sum_{e=1}^{E}\sum_{n=1}^{N}\rho_{n,e}}{\Gamma^{*}},
\end{align}
where $\Gamma_{k}^{q,e,j} = |\mathcal{N}_{k}^{q,e,j}|$ is the number of local models containing parameters $j$ in the $e$th edge communication round and $\nabla F_{n}^{j}(\bm{w}_{k,n}^{q,e,t-1})$ is the gradient of the $j$th weight.

$\mathit{Proof:}$
\begin{align}
    &\mathbb{E}\left\|\sum_{e=1}^{E}\frac{1}{\Gamma_{k}^{q,e,j}}\sum_{t=1}^{T}\sum_{n\in\mathcal{N}_{k}^{q,e,j}}[\nabla F_{n}^{j}(\bm{w}_{k,n}^{q,e,t-1}) - \nabla F_{n}^{j}(\bm{w}_{k,n}^{q,e})]\right\|^2\nonumber\\
    &\leq ET\sum_{e=1}^{E}\frac{1}{\Gamma_{k}^{q,e,j}}\!\!\sum_{t=1}^{T}\!\sum_{n\in\mathcal{N}_{k}^{q,e,j}}\!\!\!\!\!\!\mathbb{E}\|\nabla F_{n}^{j}(\bm{w}_{k,n}^{q,e,t-1})\! -\! \nabla F_{n}^{j}(\bm{w}_{k,n}^{q,e})\|^2\nonumber\\
    &\leq\frac{ET}{\Gamma^{*}}\sum_{e=1}^{E}\sum_{t=1}^{T}\sum_{n=1}^{N}\mathbb{E}\|\nabla F_{n}^{j}(\bm{w}_{k,n}^{q,e,t-1}) - \nabla F_{n}^{j}(\bm{w}_{k,n}^{q,e})\|^2\nonumber\\
    &\leq\frac{ET}{\Gamma^{*}}\sum_{e=1}^{E}\sum_{t=1}^{T}\sum_{n=1}^{N}\mathbb{E}\|\nabla F_{n}(\bm{w}_{k,n}^{q,e,t-1}) - \nabla F_{n}(\bm{w}_{k,n}^{q,e})\|^2\nonumber\\
    &\leq \frac{ET}{\Gamma^{*}}\sum_{e=1}^{E}\sum_{t=1}^{T}\sum_{n=1}^{N}L^2\mathbb{E}\|\bm{w}_{k,n}^{q,e,t-1} - \bm{w}_{k,n}^{q,e}\|^2,
\end{align}
where we relax the inequality by selecting the smallest $\Gamma^{*} = \min\Gamma_{k}^{q,e,j}$ and changing the summation over $n$ to all devices in the second step. Then, in the third step, we consider that $l_2$-gradient norm of a vector is no larger than the sum of norm of all sub-vectors, which allows us to consider $\nabla F_{n}$ rather than its sub-vectors. The last step in (45) is obtained from $L$-smoothness in Assumption 1, which ends the proof of Lemma 3.

\textbf{Lemma 4:} For IID data distribution under Assumption 4, for any global and communication rounds $q$ and $e$, we obtain that
\begin{align}
    &\!\mathbb{E}\!\left\|\sum_{e=1}^{E}\!\!\frac{1}{\Gamma_{k}^{q,e,j}}\!\!\sum_{t=1}^{T}\!\!\sum_{n\in\mathcal{N}_{k}^{q,e,j}}\!\!\!\!\![\nabla F_{n}^{j}(\bm{w}_{k,n}^{q,e,t-1}\!\!\!, \xi_{k,n}^{q,e,t-1}) \!\!- \!\!\nabla F_{n}^{j}(\bm{w}_{k,n}^{q,e,t-1})]\right\|^2\leq\frac{E^2T^2N\hat{\sigma}^2}{\Gamma^{*}}.
\end{align}

$\mathit{Proof:}$
\begin{align}
    &\!\mathbb{E}\left\|\sum_{e=1}^{E}\frac{1}{\Gamma_{k}^{q,e,j}}\sum_{t=1}^{T}\sum_{n\in\mathcal{N}_{k}^{q,e,j}}[\nabla F_{n}^{j}(\bm{w}_{k,n}^{q,e,t-1}, \xi_{k,n}^{q,e,t-1}) - \nabla F_{n}^{j}(\bm{w}_{k,n}^{q,e,t-1})]\right\|^2\nonumber\\
    &\leq\frac{ET}{\Gamma^{*}}\sum_{e=1}^{E}\sum_{t=1}^{T}\sum_{n=1}^N\mathbb{E}\|\nabla F_{n}^{j}(\bm{w}_{k,n}^{q,e,t-1}, \xi_{k,n}^{q,e,t-1}) - \nabla F_{n}^{j}(\bm{w}_{k,n}^{q,e,t-1})\|^2\nonumber\\
    &\leq\frac{ET}{\Gamma^{*}}\!\!\sum_{e=1}^{E}\sum_{t=1}^{T}\sum_{n=1}^N\!\mathbb{E}\|\nabla F_{n}(\bm{w}_{k,n}^{q,e,t-1}, \xi_{k,n}^{q,e,t-1}) - \nabla F_{n}(\bm{w}_{k,n}^{q,e,t-1})\|^2\leq\frac{E^2T^2N\hat{\sigma}^2}{\Gamma^{*}}.
\end{align}
In the second step, we consider that $l_2$-gradient norm of a vector is no larger than the sum of norm of all sub-vectors, which allows us to consider $\nabla F_{n}$ rather than its sub-vectors. The last step in (47) is obtained from gradient noise for IID data in Assumption 4, which ends the proof of Lemma 4.

\textbf{Lemma 5:} The upperbound of $\mathbb{E}\|\bm{w}_{G}^{q+1}-\bm{w}_{G}^{q}\|^2$ is denoted as
\begin{align}
    &\mathbb{E}\|\bm{w}_{G}^{q+1}-\bm{w}_{G}^{q}\|^2 \leq 3\eta^2W^2E^2T^2\phi^2 +\frac{3W^2\eta^2 E^2T^2N\hat{\sigma}^2 + 3W^2E^2L^2T^4\eta^4N\phi^2}{\Gamma^{*}}\nonumber\\
    &+\frac{6W^2\eta^2 L^2D^2T^2E\sum_{e=1}^{E}\sum_{n=1}^{N}\rho_{n,e}}{\Gamma^{*}},
\end{align}
where $W$ is the number of model weights.

$\mathit{Proof:}$
\begin{align}
    &\mathbb{E}\|\bm{w}_{G}^{q+1}-\bm{w}_{G}^{q}\|^2\nonumber\\
    &\!\!=\mathbb{E}\left\|\frac{1}{|\mathcal{K}|}\!\!\sum_{k\in\mathcal{K}}\sum_{e=1}^{E}\sum_{j=1}^{W}\frac{1}{\Gamma_{k}^{q,e,j}}\!\!\!\!\!\!\sum_{n\in\mathcal{N}_{k}^{q,e,j}}\sum_{t=1}^{T}\!\!\eta\nabla F_{n}^{j}(\bm{w}_{k,n}^{q,e,t-1}, \xi_{k,n}^{q,e,t-1})\right\|^2\nonumber\\
    &\leq \!\!\frac{3W}{|\mathcal{K}|}\sum_{k\in\mathcal{K}}\!\sum_{j=1}^{W}\!\mathbb{E}\!\left\|\sum_{e=1}^{E}\frac{1}{\Gamma_{k}^{q,e,j}}\!\!\!\sum_{n\in\mathcal{N}_{k}^{q,e,j}}\!\sum_{t=1}^{T}\!\!\eta[\nabla F_{n}^{j}(\bm{w}_{k,n}^{q,e,t-1}, \xi_{k,n}^{q,e,t-1})- \nabla F_{n}^{j}(\bm{w}_{k,n}^{q,e,t-1})]\right\|^2\nonumber\\
    &+\frac{3W}{|\mathcal{K}|}\sum_{k\in\mathcal{K}}\sum_{j=1}^{W}\mathbb{E}\left\|\sum_{e=1}^{E}\frac{1}{\Gamma_{k}^{q,e,j}}\sum_{n\in\mathcal{N}_{k}^{q,e,j}}\sum_{t=1}^{T}\eta[\nabla F_{n}^{j}(\bm{w}_{k,n}^{q,e,t-1}) - \nabla F_{n}^{j}(\bm{w}_{k,n}^{q,e})]\right\|^2\nonumber\\
    &+\frac{3W}{|\mathcal{K}|}\sum_{k\in\mathcal{K}}\sum_{j=1}^{W}\mathbb{E}\left\|\sum_{e=1}^{E}\frac{1}{\Gamma_{k}^{q,e,j}}\sum_{n\in\mathcal{N}_{k}^{q,e,j}}\sum_{t=1}^{T}\eta[\nabla F_{n}^{j}(\bm{w}_{k,n}^{q,e})]\right\|^2,
\end{align}
where we split stochastic gradient $\nabla F_{n}^{j}(\bm{w}_{k,n}^{q,e,t-1}, \xi_{k,n}^{q,e,t-1})$ into three parts, namely, $[\nabla F_{n}^{j}(\bm{w}_{k,n}^{q,e})]$, 
$[\nabla F_{n}^{j}(\bm{w}_{k,n}^{q,e,t-1}, \xi_{k,n}^{q,e,t-1}) - \nabla F_{n}^{j}(\bm{w}_{k,n}^{q,e,t-1})]$, and $[\nabla F_{n}^{j}(\bm{w}_{k,n}^{q,e,t-1}) - \nabla F_{n}^{j}(\bm{w}_{k,n}^{q,e})]$.

The third term of the last step in (49) is derived as
\begin{align}
    &\frac{3W}{|\mathcal{K}|}\sum_{k\in\mathcal{K}}\sum_{j=1}^{W}\mathbb{E}\left\|\sum_{e=1}^{E}\frac{1}{\Gamma_{k}^{q,e,j}}\sum_{n\in\mathcal{N}_{k}^{q,e,j}}\sum_{t=1}^{T}\eta[\nabla F_{n}^{j}(\bm{w}_{k,n}^{q,e})]\right\|^2\nonumber\\
    &\leq \frac{3\eta^2 WTE}{|\mathcal{K}|}\sum_{k\in\mathcal{K}}\sum_{j=1}^{W}\sum_{e=1}^{E}\sum_{t=1}^{T}\mathbb{E}\|\nabla F_{n}(\bm{w}_{k,n}^{q,e})\|^2\leq 3\eta^2W^2E^2T^2G^2.
\end{align}
Through plugging (44), (46), and (50) into (49), the upperbound of $\mathbb{E}\|\bm{w}_{G}^{q+1}-\bm{w}_{G}^{q}\|^2$ is derived as (48), which ends the proof of Lemma 5. 

\textbf{Proof of the Convergence: } Based on Lemma 2, 3, 4, and 5, we use $L$-smoothness in Assumption 1 to give convergence analysis. We begin with
\begin{equation}
    F(\bm{w}_{G}^{q+1}) \!\!\leq \!\!F(\bm{w}_{G}^{q}) + \langle\nabla F(\bm{w}_{G}^{q}), \bm{w}_{G}^{q+1}-\bm{w}_{G}^{q}\rangle + \frac{L}{2}\|\bm{w}_{G}^{q+1}-\bm{w}_{G}^{q}\|^2\!\!.
\end{equation}
Then, by taking expectations on both sides of (51), we obtain
\begin{align}
    \mathbb{E}[F(\bm{w}_{G}^{q+1})] - \mathbb{E}[F(\bm{w}_{G}^{q})] \leq\mathbb{E}\langle\nabla F(\bm{w}_{G}^{q}), \bm{w}_{G}^{q+1}-\bm{w}_{G}^{q}\rangle + \frac{
    L}{2}\mathbb{E}\|\bm{w}_{G}^{q+1}-\bm{w}_{G}^{q}\|^2.
\end{align}
First, we analyze $\mathbb{E}\langle\nabla F(\bm{w}_{G}^{q}), \bm{w}_{G}^{q+1}-\bm{w}_{G}^{q}\rangle$ by considering a sum of inner products over all model weights, which is denoted as
\begin{align}
    &\mathbb{E}\langle\nabla F(\bm{w}_{G}^{q}), \bm{w}_{G}^{q+1}-\bm{w}_{q}\rangle = \sum_{j=1}^{W}\mathbb{E}\langle\nabla F^{j}(\bm{w}_{G}^{q}), \bm{w}_{G}^{q+1,j}-\bm{w}_{G}^{q,j}\rangle\nonumber\\
    &=\sum_{j=1}^{W}\!\mathbb{E}\left\langle\nabla F^{j}(\bm{w}_{G}^{q}),-\frac{1}{|\mathcal{K}|}\sum_{k\in\mathcal{K}}\sum_{e=1}^{E}\frac{1}{\Gamma_{k}^{q,e,j}}\sum_{n\in\mathcal{N}_{k}^{q,e,j}}\sum_{t=1}^{T}\eta\nabla F_{n}^{j}(\bm{w}_{k,n}^{q,e,t-1})\right\rangle\nonumber\\
    &= - \sum_{j=1}^{W}\mathbb{E}\langle\nabla F^{j}(\bm{w}_{G}^{q}), \eta ET\nabla F^{j}(\bm{w}_{G}^{q})\rangle\nonumber\\
    &- \sum_{j=1}^{W}\mathbb{E}\left\langle\nabla F^{j}(\bm{w}_{G}^{q}),
    \frac{1}{|\mathcal{K}|}\sum_{k\in\mathcal{K}}\sum_{e=1}^{E}\frac{1}{\Gamma_{k}^{q,e,j}}\sum_{n\in\mathcal{N}_{k}^{q,e,j}}\sum_{t=1}^{T}\eta\!\left[\nabla F_{n}^{j}(\bm{w}_{k,n}^{q,e,t-1}) - \nabla F_{n}^{j}(\bm{w}_{k,n}^{q,e})\right]\right\rangle,
\end{align}
where the last step splits the result into two parts with respect to a reference point $\eta ET\nabla F^{j}(\bm{w}_{k,n}^{q,e})$. For the first term in the last step of (53), it is derived as
\begin{equation}
    - \sum_{j=1}^{W}\mathbb{E}\langle\nabla F^{j}(\bm{w}_{G}^{q}), \eta ET\nabla F^{j}(\bm{w}_{G}^{q})\rangle = -\eta ET\sum_{j=1}^{W}\left\|\nabla F^{j}(\bm{w}_{G}^{q})\right\|^2.
\end{equation}
For the second term in the last step of (53), it is derived as
\begin{align}
    &- \sum_{j=1}^{W}\mathbb{E}\left\langle\nabla F^{j}(\bm{w}_{G}^{q}),\frac{1}{|\mathcal{K}|}\sum_{k\in\mathcal{K}}\sum_{e=1}^{E}\frac{1}{\Gamma_{k}^{q,e,j}}\!\!\!\!\!\sum_{n\in\mathcal{N}_{k}^{q,e,j}}\!\sum_{t=1}^{T}\!\!\eta\!\left[\nabla F_{n}^{j}(\bm{w}_{k,n}^{q,e,t-1})\!\! -\!\! \nabla F_{n}^{j}(\bm{w}_{k,n}^{q,e})\right]\!\right\rangle\nonumber\\
    &= - \sum_{j=1}^{W}\eta ET\mathbb{E}\left\langle\nabla F^{j}(\bm{w}_{G}^{q}),\frac{1}{|\mathcal{K}|ET}\sum_{k\in\mathcal{K}}\sum_{e=1}^{E}\!\frac{1}{\Gamma_{k}^{q,e,j}}\sum_{n\in\mathcal{N}_{k}^{q,e,j}}\!\sum_{t=1}^{T}\left[\nabla F_{n}^{j}(\bm{w}_{k,n}^{q,e,t-1})\!\! -\!\! \nabla F_{n}^{j}(\bm{w}_{k,n}^{q,e})\right]\!\!\right\rangle\nonumber\\
    &\leq\frac{\eta ET}{2}\!\!\sum_{j=1}^{W}\!\mathbb{E}\|\nabla F^{j}(\bm{w}_{G}^{q})\|^2 \!\!+\!\! \frac{\eta}{2ET|\mathcal{K}|}\!\sum_{k\in\mathcal{K}}\sum_{j=1}^{W}\!\mathbb{E}\!\left\|\sum_{e=1}^{E}\frac{1}{\Gamma_{k}^{q,e,j}}\!\!\!\!\!\!\sum_{n\in\mathcal{N}_{k}^{q,e,j}}\!\sum_{t=1}^{T}\!\left[\nabla F_{n}^{j}(\bm{w}_{k,n}^{q,e,t-1})\!\! -\!\! \nabla F_{n}^{j}(\bm{w}_{k,n}^{q,e})\right]\right\|^2\nonumber\\
    &\leq\frac{\eta ET}{2}\sum_{j=1}^{W}\mathbb{E}\|\nabla F^{j}(\bm{w}_{G}^{q})\|^2 +\frac{W\phi^2NE\eta^3L^2T^3 + 2W\eta TL^2D^2\sum_{e=1}^{E}\sum_{n=1}^{N}\rho_{n,e}}{2\Gamma^{*}},
\end{align}
where the second step is obtained from (38) and the last step is obtained from Lemma 3.

By plugging (54) and (55) into (53), $\mathbb{E}\langle\nabla F(\bm{w}_{G}^{q}), \bm{w}_{G}^{q+1}-\bm{w}_{G}^{q}\rangle$ is derived as
\begin{align}
    &\mathbb{E}\langle\nabla F(\bm{w}_{G}^{q}), \bm{w}_{G}^{q+1}-\bm{w}_{G}^{q}\rangle\leq -\frac{\eta ET}{2}\sum_{j=1}^{W}\left\|\nabla F^{j}(\bm{w}_{G}^{q})\right\|^2\nonumber\\
    &+\frac{W\phi^2NE\eta^3L^2T^3 + 2W\eta TL^2D^2\sum_{e=1}^{E}\sum_{n=1}^{N}\rho_{n,e}}{2\Gamma^{*}}.
\end{align}

Finally, we plug the upperbound of  $\mathbb{E}\|\bm{w}_{G}^{q+1}-\bm{w}_{G}^{q}\|^2$ into (52) to obtain the convergence upperbound. First, we take the sum over global communication round $q = 1, 2,...,Q$ on both sides of (52) and obtain that
\begin{align}
    &\mathbb{E}[F(\bm{w}^{0})] -\!\! \mathbb{E}[F(\bm{w}^{*})] = \sum_{q=1}^{Q}\mathbb{E}[F(\bm{w}_{G}^{q+1})] \!\!- \!\!\sum_{q=1}^{Q}\mathbb{E}[F(\bm{w}_{G}^{q})]\nonumber\\
    &\leq \!\!-\sum_{q=1}^{Q}\sum_{j=1}^{W}\frac{\eta ET}{2}\left\|\nabla F^{j}(\bm{w}_{G}^{q})\right\|^2 + \sum_{q=1}^{Q}\frac{L}{2}\mathbb{E}\|\bm{w}_{G}^{q+1}-\bm{w}_{G}^{q}\|^2\nonumber\\
    &+ \sum_{q=1}^{Q}\frac{W\phi^2NE\eta^3L^2T^3 + 2W\eta TL^2D^2\sum_{e=1}^{E}\sum_{n=1}^{N}\rho_{n,e}}{2\Gamma^{*}}.
\end{align}
By plugging (48) into (57), we obtain

\begin{align}
    &\frac{\eta ET}{2}\sum_{j=1}^{W}\sum_{q=1}^{Q}\left\|\nabla F^{j}(\bm{w}_{G}^{q})\right\|^2 \leq \mathbb{E}[F(\bm{w}^{0})] - \mathbb{E}[F(\bm{w}^{*})]\nonumber\\
    &+ \frac{3QLW^2\eta^2E^2T^2\phi^2}{2}+\frac{(3QW^2\eta^2 L^3T^2D^2E + QW\eta TL^2D^2)\sum_{e=1}^{E}\sum_{n=1}^{N}\rho_{n,e}} {\Gamma^{*}} \nonumber\\
    &+\frac{3QW^2E^2L^3T^4\eta^4\phi^2N + W\phi^2NEQ\eta^3L^3T^3}{2\Gamma^{*}}+ \frac{3QLW^2\eta^2E^2T^2N\hat{\sigma}^2}{2\Gamma^{*}},
\end{align}
which completes the proof of Theorem 1.

\subsection{Appendix B - Proof of Theorem 2}
According to (6) and (30), pruning ratio $\rho_{n,e}$ is calculated as
\begin{align}
&\left(W_{n,\text{conv}} + (1 - \rho_{n,e})W_{n,\text{fully}}\right)\left(\frac{TC_n}{f_n} + \frac{\hat{q}}{R_{n,k,e}^{\text{up}}}\right) \leq T_{\text{th}}, \\
&T_{n,e}^{\text{cmp-conv}} + T_{n,k,e}^{\text{com-conv}} + (1 - \rho_{n,e})(T_{n,e}^{\text{cmp-fully}} + T_{n,k,e}^{\text{com-fully}}) \leq T_{\text{th}},
\end{align}
where $T_{n,e}^{\text{cmp-conv}} = W_{n,\text{conv}}TC_n/f_n$, $T_{n,k,e}^{\text{com-conv}} = W_{n,\text{conv}}\hat{q}/R_{n,k,q}^{\text{up}}$, 
$T_{n,e}^{\text{cmp-fully}} = W_{n,\text{fully}}TC_n/f_n$, and $T_{n,k,e}^{\text{com-fully}} = W_{n,\text{com-fully}}\hat{q}/R_{n,k,e}^{\text{up}}$. Then, $\rho_{n,e}$ is deduced as (31), which ends the proof of Theorem 2.

\subsection{Appendix C - Proof of Lemma 1}
The objective function in (33) is equal to 
\begin{equation}
    F(X) = \sum_{e=1}^{E}\sum_{n=1}^{N}f(x_{n,e}) = \sum_{e=1}^{E}\sum_{n=1}^{N}\left(1 - \frac{x_{n,e}V_1 - V_2}{x_{n,e}V_3 + V_4}\right),
\end{equation}
where $V_1, V_2, V_3, V_4 > 0$, and $0\leq x_{n,e}\leq 1$. To prove the lemma, we just need to analyze the convexity of the function $f(x_{n,e})$. The first derivative is derived as
\begin{equation}
    f^{'}(x_{n,e}) = -\frac{V_1V_4 + V_2V_3}{(V_3x_{n,e} + V_4)^2}.
\end{equation}
Then, the second derivative is calculated as
\begin{equation}
    f^{''}(x_{n,e}) = \frac{2V_3(V_1V_4 + V_2V_3)(V_3x_{n,e} + V_4)}{(V_3x_{n,e} + V_4)^4}\geq 0.
\end{equation}
Therefore, the objective function in (33) is convex. Also, both constraints in (27) and (28) are convex. As a result, the optimization problem in (33) in convex, which ends the proof of Lemma 1.

\subsection{Appendix D - Proof of Theorem 3}
Based on the optimization in (33) and constraint (27), the Lagrange function is denoted as
\begin{align}
    \mathcal{L}({b}_{n,e}, \lambda) \!= \!H_2\!\sum_{e=1}^{E}\sum_{n=1}^{N}\!\!\left(\!\!1 \!- \!\frac{R_{n,k,e}^{\text{up}}(T_{\text{th}} - T_{n,e}^{\text{cmp-conv}}) - \hat{q}W_{n,\text{conv}}}{R_{n,k,e}^{\text{up}}T_{n,e}^{\text{cmp-fully}} + \hat{q}W_{n,\text{fully}}}\right)+ \lambda\left(\sum_{n=1}^{N}{b}_{n,e} - 1\right),
\end{align}
where $\lambda$ is the Lagrange multiplier. Then, the Karush-Kuhn-Tucker (KKT) conditions are deduced as
\begin{align}
    \frac{\partial\mathcal{L}}{\partial {b}_{n,e}} =\lambda - \frac{(V_1V_4+V_2V_3)B\log_{2}\left(1 + \frac{g_{n,k}^{e}p_n}{\sigma^2}\right)}{\left({b}_{n,e}BV_3\log_{2}\left(1 + \frac{g_{n,k}^{e}p_n}{\sigma^2}\right) + V_4\right)^2}= 0,
\end{align}
\begin{equation}
    \lambda\left(\sum_{n\in\mathcal{N}_k}{b}_{n,e} - 1\right) = 0,
\end{equation}
\begin{equation}
    \lambda\geq 0.
\end{equation}

Based on the KKT conditions, the optimal bandwidth allocation is achieved as Theorem 3, which ends the proof of Theorem 3.

%





\ifCLASSOPTIONcaptionsoff
  \newpage
\fi





\bibliographystyle{IEEEtran}
\bibliography{IEEEabrv,Ref,ReferencesMP}
%
%

\end{document}